\documentclass[10pt,twocolumn,letterpaper]{article}
\usepackage[accsupp]{axessibility}  
\usepackage[pagenumbers]{cvpr} 

\renewcommand{\paragraph}[1]{\vspace{.5em}\noindent\textbf{#1}}

\usepackage{graphicx}
\usepackage{cuted}
\usepackage{caption,subcaption}
\usepackage{booktabs} 
\usepackage{tabularx}
\usepackage{pifont}
\usepackage{multirow}
\usepackage[dvipsnames]{xcolor}
\usepackage{color, colortbl}
\usepackage{svg}
\usepackage{epstopdf}
\usepackage{tikz}
\usepackage{stfloats,float}
\usepackage{indentfirst}
\usepackage{algorithm, algpseudocode}
\usepackage{listings}

\definecolor{Lblue}{rgb}{0.69,0.92,0.95}
\definecolor{gray}{rgb}{0.7,0.7,0.7}
\definecolor{lightgray}{rgb}{0.6,0.6,0.6}

\usepackage{amsmath,amsfonts,bm}

\def\eqref#1{equation~\ref{#1}}

\def\1{\bm{1}}

\def\rvx{{\mathbf{x}}}

\DeclareMathAlphabet{\mathsfit}{\encodingdefault}{\sfdefault}{m}{sl}
\SetMathAlphabet{\mathsfit}{bold}{\encodingdefault}{\sfdefault}{bx}{n}

\graphicspath{{./figures/}}

\newcommand{\CIFARBIFMFIDEpochs}{%
(50,  6.8530)
(100, 5.5930)
(150, 4.3114)
(200, 3.6996)
(250, 3.1992)
(300, 2.9713)
(350, 2.8029)
(400, 2.7319)
(450, 2.6294)
(500, 2.4737)
}

\newcommand{\CIFARFlowMatchingFIDEpochs}{%
(50,  7.3345)
(100, 5.9160)
(150, 4.6644)
(200, 3.9787)
(250, 3.4831)
(300, 3.1621)
(350, 2.9219)
(400, 2.8594)
(450, 2.7545)
(500, 2.5648)
}

\newcommand{\CIFARFlowMatchingKeyPoint}{%
(250, 3.4831)
}

\newcommand{\CIFARBiFMKeyPoint}{%
(250, 3.1992)
}

\newcommand{\figTrainingBenefitFID}{%
\renewcommand{\hh}{87mm}\renewcommand{\vv}{57mm}\renewcommand{\hhh}{-5mm}%
\begin{figure}[h]
\centering%
\resizebox{\linewidth}{!}{%
\hspace*{-1mm}%
\begin{tikzpicture}%
\begin{axis}[
  width=\hh, height=\vv,
  xmin={0}, xmax={500}, xtick={0, 100, 200, 300, 400, 500}, xticklabels={{\hspace{2.5em}Epoch}, $100$, $200$, $300$, $400$, $500$},
  ymin={2}, ymax={7}, ymode=log, ytick={2, 3, 4, 5, 6, 7}, yticklabels={$2.0$, $3.0$, $4.0$, $5.0$, $6.0$, {FID}},
  grid={major}, legend pos={north east}, legend cell align={left}, legend style={font=\small},
]
\addplot[C1]                             coordinates {\CIFARFlowMatchingFIDEpochs};
\addplot[C0]                             coordinates {\CIFARBIFMFIDEpochs};
\fillbetween[C0, opacity=0.10, forget plot]{coordinates {\CIFARBIFMFIDEpochs}}{coordinates {\CIFARFlowMatchingFIDEpochs}};
\addplot[C0, only marks]        coordinates {\CIFARBiFMKeyPoint};
\addplot[C1, only marks]        coordinates {\CIFARFlowMatchingKeyPoint};
\addplot[dashed, thin] coordinates {(250,2.0) (250,3.48) };
\node at (axis cs:280,4.0) [anchor={north}] {\textcolor{C1}{$3.48$}};
\node at (axis cs:280,2.9) [anchor={north}] {\textcolor{C0}{$3.20$}};
\legend{
  {FM~\citep{lipman2023fm}},
  {FM + BiFM (ours)},
}
\end{axis}
\end{tikzpicture}
}
\vspace*{-7mm}
\caption{\label{fig:cifarbenefit}%
{\bf CIFAR-10 Training Epochs vs. FID}}
\end{figure}
}

\definecolor{olive}{rgb}{0.5, 0.5, 0.0}
\definecolor{maroon}{rgb}{0.69, 0.19, 0.38}
\definecolor{celestialblue}{rgb}{0.29, 0.59, 0.82}
\definecolor{darkgreen}{rgb}{0.0, 0.6, 0.0}
\definecolor{grey}{rgb}{0.5,0.5,0.5}
\definecolor{darkblue}{rgb}{0.19, 0.19, 0.62}
\definecolor{silver}{rgb}{0.7,0.7,0.7}
\definecolor{darkcyan}{rgb}{0.0, 0.55, 0.55}

\def\clap#1{\hbox to 0pt{\hss #1\hss}}%

\newcommand\undefcolumntype[1]{\expandafter\let\csname NC@find@#1\endcsname\relax}

\usepackage{pgfplots}
\pgfplotsset{compat=1.18}
\usepgfplotslibrary{fillbetween}
\pgfplotsset{xtick style={draw=none}}
\pgfplotsset{ytick style={draw=none}}
\pgfplotsset{major grid style={gray!40}}
\pgfplotsset{every axis plot/.style={thick, mark size=1.5pt}}
\pgfplotsset{legend image code/.code={\draw[mark repeat=2, mark phase=2] plot coordinates {(0cm, 0cm) (0.2cm, 0cm) (0.4cm, 0cm)};}} %

\definecolor{C0}{rgb}{0.121569, 0.466667, 0.705882}
\definecolor{C1}{rgb}{1.000000, 0.498039, 0.054902}
\definecolor{C2}{rgb}{0.172549, 0.627451, 0.172549}
\definecolor{C3}{rgb}{0.839216, 0.152941, 0.156863}
\definecolor{C4}{rgb}{0.580392, 0.403922, 0.741176}
\definecolor{C5}{rgb}{0.549020, 0.337255, 0.294118}
\definecolor{C6}{rgb}{0.890196, 0.466667, 0.760784}
\definecolor{C7}{rgb}{0.498039, 0.498039, 0.498039}
\definecolor{C8}{rgb}{0.737255, 0.741176, 0.133333}
\definecolor{C9}{rgb}{0.090196, 0.745098, 0.811765}

\newcommand{\fillbetween}[3][]{\addplot+[name path=A, draw=none, mark=none, forget plot] #2; \addplot+[name path=B, draw=none, mark=none, forget plot] #3; \addplot[#1] fill between[of=A and B]}

\newcommand{\hh}{0mm}
\newcommand{\hhh}{0mm}

\newcommand{\vv}{0mm}

\definecolor{cvprblue}{rgb}{0.21,0.49,0.74}
\usepackage[pagebackref,breaklinks,colorlinks,allcolors=cvprblue]{hyperref}

\def\paperID{46451} 
\def\confName{CVPR}
\def\confYear{2026}

\title{BiFM: Bidirectional Flow Matching for Few-Step Image Editing and Generation}

\author{
Yasong Dai\textsuperscript{\rm 1, 2},$\;$
Zeeshan Hayder\textsuperscript{\rm 1, 2},$\;$
David Ahmedt-Aristizabal\textsuperscript{\rm 2},$\;$
Hongdong Li\textsuperscript{\rm 1}
\\
\textsuperscript{\rm 1}Australian National University, Australia,$\;$
\textsuperscript{\rm 2}Data61-CSIRO, Australia
\\
{\tt\small \{yasong.dai, hongdong.li\}@anu.edu.au}\\
{\tt\small \{zeeshan.hayder, david.ahmedtaristizabal\}@data61.csiro.au}
}

\begin{document}
\maketitle

\begin{abstract}

Recent diffusion and flow matching models have demonstrated strong capabilities in image generation and editing by progressively removing noise through iterative sampling. While this enables flexible inversion for semantic-preserving edits, few-step sampling regimes suffer from poor forward process approximation, leading to degraded editing quality. Existing few-step inversion methods often rely on pretrained generators and auxiliary modules, limiting scalability and generalization across different architectures.
To address these limitations, we propose BiFM (Bidirectional Flow Matching), a unified framework that jointly learns generation and inversion within a single model. BiFM directly estimates average velocity fields in both ``image $\to$ noise" and ``noise $\to$ image" directions, constrained by a shared instantaneous velocity field derived from either predefined schedules or pretrained multi-step diffusion models. Additionally, BiFM introduces a novel training strategy using continuous time-interval supervision, stabilized by a bidirectional consistency objective and a lightweight time-interval embedding. This bidirectional formulation also enables one-step inversion and can integrate seamlessly into popular diffusion and flow matching backbones.
Across diverse image editing and generation tasks, BiFM consistently outperforms existing few-step approaches, achieving superior performance and editability. 

\end{abstract}    
\section{Introduction}
\label{sec:introduction}


\definecolor{reblue}{HTML}{5690F1}
\definecolor{reyellow}{HTML}{FFC107}
\definecolor{rered}{HTML}{E95758}

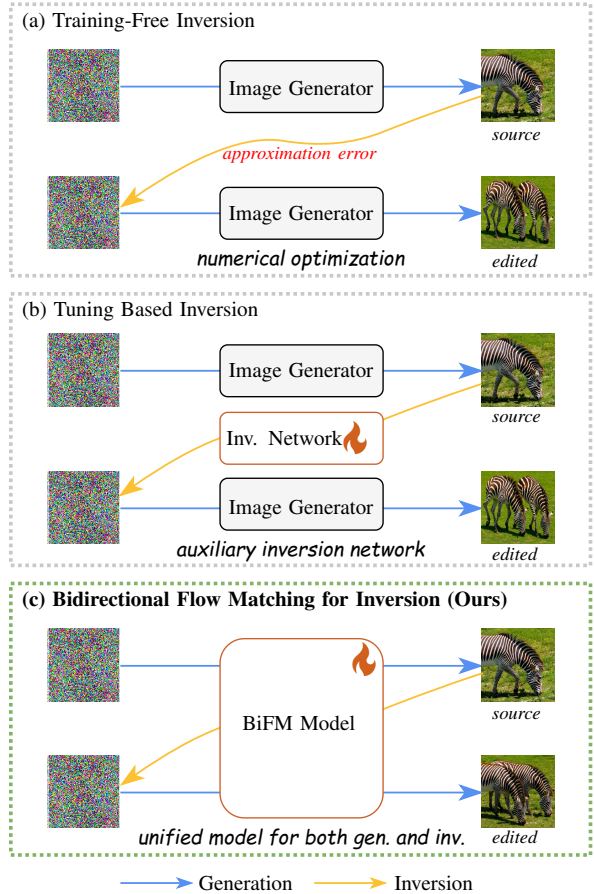
\begin{figure}[t]
    \centering
    \small
    \begin{tikzpicture}[every node/.style={ font=\footnotesize,
                                            text width=\linewidth,
                                            align=left,
                                            anchor=south west
                                            }]

    \node[anchor=south west, inner sep=0] (image) at (0,0) {\includegraphics[width=0.95\linewidth]{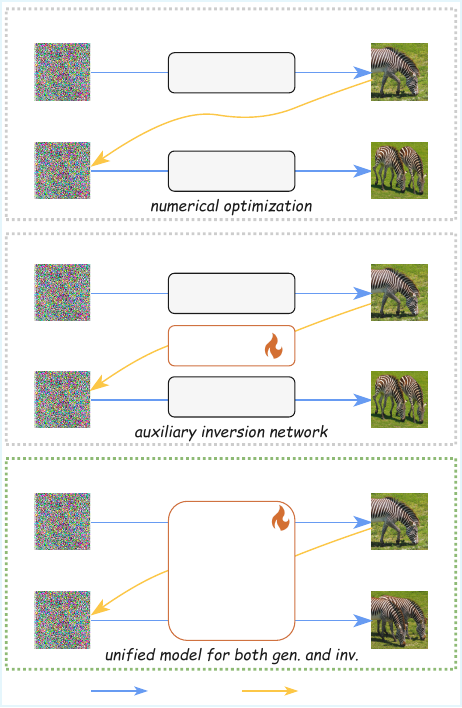}};

    \node[] at (0.15, 11.50) {(a) Training-Free Inversion};
    \node[] at (6.40, 10.05) {\emph{\scriptsize source}};
    \node[] at (6.40, 8.37) {\emph{\scriptsize edited}};
    \node[] at (2.90, 10.60) {Image Generator};
    \node[] at (2.90, 8.93) {Image Generator};
    \node[] at (2.82, 9.75) {\scriptsize\textcolor{red}{\emph{approximation error}}};

    \node[] at (0.15, 7.65) {(b) Tuning Based Inversion};
    \node[] at (6.40, 6.30) {\emph{\scriptsize source}};
    \node[] at (6.40, 4.47) {\emph{\scriptsize edited}};
    \node[] at (2.90, 6.85) {Image Generator};
    \node[] at (2.90, 5.02) {Image Generator};
    \node[] at (2.87, 6.00) {Inv. Network};

    \node[] at (0.15, 3.80) {\bf (c) Bidirectional Flow Matching for Inversion (Ours)};
    \node[] at (6.40, 2.35) {\emph{\scriptsize source}};
    \node[] at (6.40, 0.70) {\emph{\scriptsize edited}};
    \node[] at (3.08, 2.20) {BiFM Model};

    \node[] at (2.50, 0.08) {Generation};
    \node[] at (5.10, 0.08) {Inversion};    
      
    \end{tikzpicture}
    \vspace{-1.5em}  
    \caption{{\bf Inversion-Based Image Editing.} 
    (a) In training-free inversion, the process is approximated by numerically reversing the generation steps, leading to accumulated approximation errors; (b) An auxiliary inversion network is introduced on top of a pretrained generator, improving fidelity but increasing complexity and reducing generalization across architectures. (c) Our method, BiFM, jointly learns generation and inversion within a single flow matching model, enabling consistent few-step inversion and editing.
    }
    \vspace{-2em}
    \label{fig:intro_teaser}
\end{figure}

Diffusion models~\citep{ho2020ddpm,song21ddim} and flow matching models~\citep{lipman2023fm,liu2022rectifiedflow} achieve strong image generation by learning the data distribution through multi-step sampling. Their generation process can be viewed as solving a learned time-dependent probability flow ordinary differential equation (ODE)~\citep{song2021score} starting from random noise.
An important application of diffusion models is inversion-based image editing~\citep{song21ddim,mokady2023null,wang2025rfedit,garibi2024renoise}, where a source image is mapped back to the intermediate latent space of a generative model and then forwarded again with target prompts. The inversion process enables controllable and semantically faithful edits, but it is naively slow because it doubles the number of inference steps. Consequently, recent research has focused on few-step editing methods~\citep{wu2024turboedit,starodubcev2024icd,nguyen2025swiftedit}, which offer significant advantages in inference speed and enable real-time interactive editing. 

However, a key gap remains: the inversion process for few-step diffusion models is intrinsically hard to learn. Few-step models utilize large time-step updates, which amplify the approximation error from local linearization~\citep{song21ddim} and ODE solvers~\citep{lu2022dpmsolver}. 
An overview of existing inversion-based editing approaches and our proposed bidirectional framework is illustrated in \Cref{fig:intro_teaser}:
Training-free inversion process inherits approximation errors~\citep{wallace2023edict,hong2024exact,zhang2023bdia}, causing semantic drift or background preservation issues during few-step editing.
Tuning based methods~\citep{nguyen2025swiftedit,starodubcev2024icd} learn an inversion process by introducing additional inversion networks and task-specific modules, but incur additional training and computational overhead. Our core motivation arises from this gap: \emph{Can we train a few-step diffusion model that directly learns its own inversion process?}

Motivated by these observations, our research question is how few-step generation and inversion can be jointly learned and how such joint learning can improve generation and editing performance under a few-step sampling budget.
To achieve invertibility, bidirectional (invertible) neural networks are often implemented as variants of affine coupling layers~\citep{wang2024belm} that can be explicitly inverted.
In contrast, we obtain inversion in a natural way from the ODE viewpoint by integrating the flow matching ODE in both time directions.
For few-step generation, instead of learning the entire ODE trajectory, we parameterize the model as the average velocity field of the flow matching ODE over continuous time intervals.

To this end, we introduce \emph{\bf BiFM} (Bidirectional Flow Matching), a flow matching model that directly learns both few-step generation and inversion. We provide a training and sampling pipeline for BiFM that predicts bidirectional average velocities along the flow matching ODE. Following prior work~\citep{frans2025shortcut,geng2025meanflow,boffi2025flowmap}, our few-step training uses arbitrary time intervals for supervision. To enable generative training in both time directions, we establish a physically constrained connection between forward (generation) and backward (inversion) average velocity fields by extending the MeanFlow Identity proposed by \citet{geng2025meanflow}. To stabilize training, we propose a bidirectional consistency training objective and a lightweight time-interval embedding that can be seamlessly integrated into popular diffusion and flow model backbones, such as SiT and MMDiT.

BiFM offers accurate inversion and image editing and can be either fine-tuned from pretrained flow matching models or trained from scratch. Across a wide range of image editing and generation tasks, BiFM consistently outperforms previous few-step methods. Compared with training-free inversion approaches, BiFM adds minimal training cost while retaining state-of-the-art performance on image editing benchmarks. To justify key design choices, we also conduct ablation studies on one-step image generation.

In summary, our contributions are:
\begin{itemize}
    \item We propose BiFM, a joint generation and inversion flow matching framework, enabling generation and inversion-based editing under a few-step sampling budget.
    \item We demonstrate that BiFM can be applied to large pretrained text-to-image diffusion models for efficient fine-tuning on image editing tasks.
    \item We comprehensively evaluate the performance of BiFM on image editing and various image generation tasks, and provide ablation studies that clarify the impact of core design choices.
\end{itemize}
\section{Related Work}
\label{sec:relatedwork}

\subsection{Inversion based Image Editing}

\noindent\textbf{Training-free inversion.} 
DDIM inversion~\citep{song21ddim} enables editing by reversing the deterministic sampling trajectory. However, its effectiveness relies on local linearity assumptions, which degrade under large step sizes. Subsequent methods~\citep{wang2025rfedit,hong2024exact} introduce solver-specific inversion techniques to mitigate these issues, but still depend on approximate dynamics and lack robustness in few-step regimes.

\noindent\textbf{Tuning few-step/one-step generators.} 
To support interactive editing, recent approaches fine-tune or distill editors from pretrained few-step generators~\citep{starodubcev2024icd}, or attach auxiliary inversion modules~\citep{nguyen2025swiftedit,wu2024turboedit}. While these methods improve efficiency, they often inherit instability from the underlying generators and introduce additional parameters or heuristics for inversion, thus limiting generalization.

\subsection{Efficient Diffusion Distillation}

Recent approaches to accelerate sampling in diffusion and flow matching models fall into three main categories. First, non-Markovian deterministic sampling methods such as DDIM~\citep{song21ddim}. Second, progressive distillation~\citep{salimans2022progressive,frans2025shortcut} that trains compact student models to mimic teacher models. Third, time-interval supervision approaches~\citep{geng2025meanflow,wang2025tim} directly supervise integrated flow matching ODE over continuous time spans, enabling efficient few-step generation.
While the third category enables flexible and simplified training recipes, the introduction of time-interval often leads to training instability and typically not applied on inversion, which limits their applicability to inversion-based editing tasks in few sampling steps.

\subsection{Invertible Neural Networks}

Recent work has explored architectural and training strategies to enable invertibility in generative models. Some approaches impose structural constraints to enforce invertibility~\citep{wang2024belm,wallace2023edict}, while others jointly learn forward and backward mappings through consistency-based distillation~\citep{li2024bcm} or invertible consistency models~\citep{starodubcev2024icd}. For example, iCD~\citep{starodubcev2024icd} demonstrates text-guided editing with around seven steps using consistency distillation and dynamic classifier-free guidance (CFG), highlighting the importance of bidirectional control in text-to-image tasks.
While these models show promise for image editing, they have primarily been evaluated on simplified datasets and tasks. The open challenge remains whether such invertible architectures can be effectively applied to complex generative models to reduce inversion error and support high-quality editing in realistic scenarios.
\section{Rethink Inversion and Few-Step Diffusion}
\label{sec:rethink}

\subsection{Diffusion Model and Flow Matching}

Diffusion and flow matching models are generative models that learn a transformation from prior noise distribution $\mathcal{N}(0,I)$ to an unknown complex data distribution $p_\text{data}(\rvx)$.

\paragraph{Denoising diffusion models}~\citep{ho2020ddpm,nichol2021improved,song2021score} construct this transformation via a forward Markov process $q(\rvx_t|\rvx_{t-1})$ that gradually adds noise to data, and a learned reverse process $p_\theta(\rvx_{t-1}|\rvx_{t})$ that denoises it. 
Training is typically performed by maximizing the log-likelihood of $p_\theta(\rvx_0)$, which can be simplified to a denoising objective in~\cref{eq:dm_simple}, following~\citep{ho2020ddpm}
\vspace{-4pt}:
\begin{equation} \label{eq:dm_simple}
    \mathcal{L}_\text{denoise} = \mathbb{E}_{t,\rvx_0,\epsilon} \left[||\epsilon-\epsilon_\theta(\rvx_t,t)||^2\right]
\vspace{-4pt}
\end{equation}
\noindent{\bf Flow matching}~\citep{lipman2023fm,liu2022rectifiedflow} learns a time-dependent velocity field $v_\theta(\rvx_t,t),t\in[0,1],$ that drives a continuous flow from noise to data. Given $\rvx_0 \sim \mathcal{N}(0, I)$ and $\rvx_1 \sim p_\text{data}(\rvx)$, the flow path is defined as a linear interpolation: $\rvx_t := (1-t)\rvx_0 + t \rvx_1$. 
Flow matching does not rely on a predefined forward process but learns from conditional velocity $v_t | \rvx_t$ that depends on the coupling between $(\rvx_0, \rvx_1)$. An effective objective for training is the conditional flow matching loss~\citep{lipman2023fm}, as in~\cref{eq:cfm_loss}:
\begin{equation}\label{eq:cfm_loss}
    \mathcal{L}_\text{CFM} = \mathbb{E}_{t,\rvx_0,\rvx_1} \left[||v_\theta(\rvx_t,t)-v_t(\rvx_t|\rvx_0,\rvx_1)||^2\right]
\end{equation}
At inference, sampling reduces to solving the learned probability flow ODE $d \rvx_t / dt = v_\theta(\rvx_t,t)$, starting from $\rvx_0$.

\paragraph{Time convention.}
Below, we use the flow matching time convention $t \in [0,1]$ (noise $\to$ data). When relating to diffusion models we treat the discrete diffusion time index as a reparameterization of $t$.

\subsection{Limitations of DDIM-based Inversion}
Inversion~\citep{chen2025inversionsurvey} refers to the process of mapping an image back to its intermediate latent within a pretrained generative model. DDIM~\citep{song21ddim} inversion is widely used due to its deterministic formulation, which simplifies the sampling process by zeroing ${\sigma_t^2}$, resulting in the following generation step:
\begin{equation}
    \tilde{\rvx}_{t+\Delta t} = \tilde{\rvx}_t +  \delta_t\epsilon_\theta(\rvx_t,t)
\label{eq:ddim_forward}
\end{equation}
where $\tilde{\rvx}_t:=\rvx_t/\sqrt{\alpha_t}$, $\delta_t := \sqrt{\frac{1-\alpha_{t-1}}{\alpha_{t-1}}} - \sqrt{\frac{1-\alpha_t}{\alpha_t}}$. DDIM allows deterministic inversion by reversing \cref{eq:ddim_forward}:
\begin{equation}
        \tilde{\rvx}_{t} = \tilde{\rvx}_{t+\Delta t} - \delta_t\epsilon_\theta(\rvx_{t},t) \approx \tilde{\rvx}_{t+\Delta t} -  \delta_t {\epsilon_\theta(\rvx_{t+\Delta t},t)} 
\label{eq:ddim}
\end{equation}
This training-free inversion mechanism is appealing for image editing tasks. However, in the few-step regime, the discrepancy between consecutive noise predictions, $|\epsilon_\theta(\rvx_t,t) -\epsilon_\theta(\rvx_{t+\Delta t},t)|$, becomes significant due to large step sizes. As illustrated in \Cref{fig:rethink_ddim}~(c), this leads to poor latent recovery and degraded editing quality, making DDIM inversion unreliable for few-step applications.

\subsection{Time-Interval Distillation of Diffusion Models}

The need for iterative sampling via ODE/SDE solvers limits efficiency for diffusion and flow models, especially in few-step regimes where speed and fidelity are critical.
Similar principle applies to flow matching ODE. By leveraging the physical definitions of average and instantaneous velocity, we define average velocity field $u(\rvx_t,t,t')$ as the integral of the instantaneous velocity $v(\rvx_t,t)$ over a time interval $[t,t']$:
\vspace{-2pt}
\begin{equation}\label{eq:mean_velocity}
    u(\rvx_{t},t,t') := \frac{1}{t'-t}\int_{t}^{t'} v(\rvx_\tau,\tau) d\tau
\end{equation}
\cref{eq:mean_velocity} enables few-step training by supervising the model to match average velocities over time intervals, rather than relying on dense trajectory sampling. However, existing flow matching models typically focus on generation without addressing inversion or editing tasks. In the next section, we introduce BiFM, a bidirectional framework that unifies generation and inversion under a shared velocity field.

\section{Bidirectional Flow Matching (BiFM)}
\label{sec:method}

\begin{figure*}
    \centering
    \includegraphics[width=0.95\linewidth]{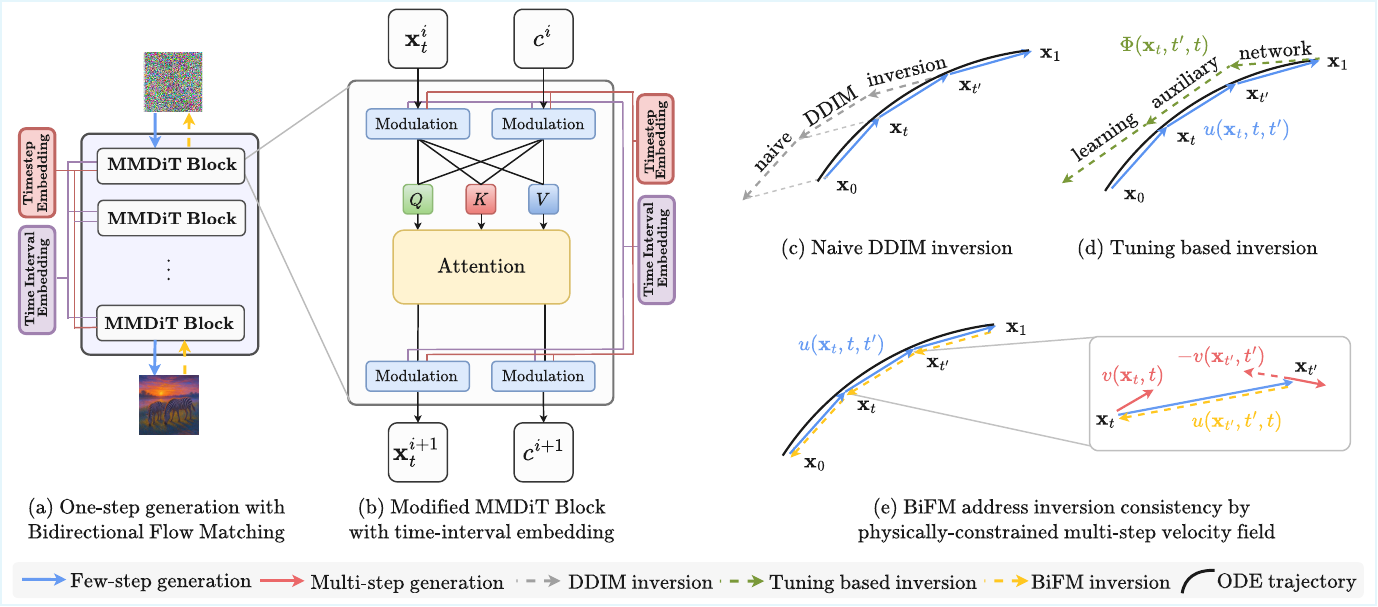}
    \caption{{\bf Overview of BiFM.}
    (a) Our one-step generation architecture built upon MMDiT based flow matching model.
    (b) A single MMDiT block showing how time embedding modulation impact model output.
    (c) Naive DDIM inversion reuses DDIM update in reverse time, causing departures from original ODE trajectory in few-step regime.
    (d) Tuning based inversion introduces an auxiliary network $\Phi(\rvx_{t'},t',t)$
    (e) BiFM inversion (ours) learns a physically constrained bidirectional average velocity field.
    }
    \label{fig:rethink_ddim}
\end{figure*}

In this section, we introduce the core components of BiFM, our proposed framework for jointly learning few-step generation and inversion.
Building on flow matching and time-interval supervision, BiFM extends the MeanFlow Identity to support the modeling of flow matching ODE in both time directions within a single model. Our key insight is that both forward and backward average velocity fields are defined with respect to a shared instantaneous velocity field. This allows BiFM to perform high-fidelity few-step sampling and inversion without relying on DDIM-based inversion or auxiliary modules.

\subsection{MeanFlow Identity for One-Step Generator}
\label{subsec:meanflow}

We start by revisiting the MeanFlow Identity~\citep{geng2025meanflow} that links the average velocity field $u(\rvx_t, t, t')$ to the instantaneous velocity $v(\rvx_t, t)$ in flow matching ODE.
Recall the average velocity over a time interval $[t, t']$:
\begin{equation}
    u(\rvx_t, t, t') 
    := \frac{1}{t' - t} \int_{t}^{t'} v(x_\tau, \tau)\, d\tau.
    \label{eq:average_v}
\vspace{-2pt}
\end{equation}
Assuming $t$ and $t'$ are independent, differentiating \cref{eq:average_v} with respect to $t$ yields \cref{eq:mf_diff}, where the total derivative can be expressed using the Jacobian-vector product (JVP) as shown in \cref{eq:mf_jvp}:
\vspace{-3pt}
\begin{gather}
    u(\rvx_{t}, t, t') = v(\rvx_{t}, t) + (t'-t)\cdot\frac{d}{dt} u(\rvx_{t}, t, t') \label{eq:mf_diff} \\
    \frac{d}{dt} u(\rvx_{t}, t, t') = v(\rvx_{t}, t)\partial_{\rvx_{t}}{u}+\partial_{t}{u} \label{eq:mf_jvp}
\end{gather}
Together, \cref{eq:mf_diff} and (\ref{eq:mf_jvp}) define the MeanFlow Identity that gives expression of target average velocity field $u_\text{tgt}$ to be computed during training:
\begin{equation}
    u_{\text{tgt}}=v(\rvx_t,t)+(t'-t)\cdot\left[v(\rvx_t,t)\partial_{\rvx_t}u_\theta+\partial_{t}u_\theta\right] \label{eq:mf_id}
\end{equation}
By introducing \cref{eq:mf_id}, we can train a few-step generation model either from scratch (where $v(\rvx_t,t)$ is pre-defined schedule like rectified flow~\citep{liu2022rectifiedflow}) or by fine-tuning (where $v(\rvx_t,t)$$:=v_\theta(\rvx_t,t)$ is a pretrained multi-step generator). The training objective $\mathcal{L}_\text{MF}$ regresses a parameterized average velocity field $u_\theta$ towards $u_{\text{tgt}}$:
\begin{equation}
    \mathcal{L}_{\text{MF}} = \mathbb{E}_{t,t',\rvx}\left[|| u_\theta(\rvx_t,t,t') -\text{sg}(u_{\text{tgt}})||^2 \right]\\ 
\end{equation}
Here $\text{sg}(\cdot)$ denotes stop-gradient operation. Intuitively, $u_\theta$ is trained so that taking one average-velocity step from $t$ towards $t'$ approximates integrating the underlying flow matching ODE over $[t, t']$. At convergence, $u_\theta$ behaves as a one-step generator consistent with the underlying multi-step dynamics. \Cref{fig:rethink_ddim}~(e) conceptually illustrates this behavior: multi-step sampling closely follows the underlying flow matching ODE trajectory, while a few-step model aims to approximate the same trajectory using large time intervals through learned average velocities.

\subsection{Extend Time Directions for Flow Inversion}
\label{subsec:bifm}

\noindent{\bf Motivation.} Most diffusion and flow matching models are trained under a fixed time convention: from noise to data. However, inversion-based editing requires the \emph{opposite} operation: from image to noise. Therefore, DDIM inversion suffers from approximation errors because the model requires $\rvx_t$ as input to compute $\epsilon_\theta(\rvx_t, t)$, yet inversion starts from $\rvx_{t+\Delta t}$, illustrated in \Cref{fig:rethink_ddim}(c). This mismatch leads to poor inversion in few-step regimes.

Our key observation is that this limitation is largely a consequence of the {time convention} enforced during training. In the remainder of this subsection,
we show how to realize this idea using a bidirectional consistency loss term derived from the MeanFlow Identity.

\noindent{\bf Bidirectional consistency objective.}
Although MeanFlow Identity typically assumes $t < t'$ for training and sampling, the identity itself does not depend on this ordering and holds for $t > t'$, allowing us to define inversion using the same formulation. Specifically, given $t, t'$ with $t < t'$, we interpret $u(\rvx_t, t, t')$ as generation (forward average velocity) and $u(\rvx_{t'}, t', t)$ as inversion (backward average velocity). Both are defined from the \emph{same} instantaneous velocity field $v(\rvx_t,t)$, but integrated over opposite time intervals. Applying \cref{eq:mf_id} gives:
\begin{equation*}
\begin{split}
& u(\rvx_t, t, t') = v(\rvx_t, t) + (t'-t)\bigl(v(\rvx_t, t)\partial_{\rvx_t}u + \partial_t u\bigr) \\
& u(\rvx_{t'}, t', t) = v(\rvx_{t'}, t') + (t-t')\bigl(v(\rvx_{t'}, t')\partial_{\rvx_{t'}}u + \partial_t u\bigr)
\end{split}
\end{equation*}
In continuous time, the average velocity for the backward interval $[t', t]$ is the negative of the forward average velocity over $[t, t']$, evaluated at corresponding points along the true trajectory. This is precisely the notion of reversibility we want in a generative model used for inversion-based editing: going forward from $(x_t, t)$ to $(x_{t'}, t')$ and then backward to $(x_t, t)$ should approximately recover the original state, as shown in \Cref{fig:rethink_ddim}(e).

To encode this reversibility at the level of the learned average velocity, we explicitly encourage the forward and backward predictions to be negatives of each other by introducing a bidirectional consistency loss:
\vspace{-3pt}
\begin{equation}
    \mathcal{L}_\text{BiFM} = \mathcal{D}\left(u_\theta(\rvx_{t}, t, t'),- u_\theta(\rvx_{t'}, t', t)\right)
    \vspace{-3pt}
\end{equation}
where $D(\cdot,\cdot)$ is a distance metric (e.g., a robust $\ell_p$ norm; see \Cref{subsec:ablation} for choices). This consistency term penalizes the discrepancy between the forward average velocity and backward average velocity. Our final training objective combines the bidirectional consistency term with $\mathcal{L}_\text{MF}$:
\begin{equation}
    \mathcal{L} = \mathcal{L}_\text{MF} + w(t,t')\cdot\mathcal{L}_\text{BiFM}
\end{equation}
Here, $w(t, t')$ is a time-dependent weighting schedule that stabilizes training by gradually strengthening the bidirectional constraint. This formulation allows BiFM to be both trained from scratch and distilled from multi-step models, enabling efficient few-step generation and inversion.

\subsection{BiFM Fine-Tuning from Pretrained Models}

While most flow matching models adopt a simple velocity field $v_t := \rvx_1 - \rvx_0$ during training, BiFM also applies to pretrained diffusion models whose $v_\theta(\rvx_t,t)$ traces more complex, curved trajectories. Importantly,~\cref{eq:mf_id} does not require explicit access to the instantaneous velocity field, allowing BiFM to be fine-tuned on pretrained models with objective $\mathcal{L}$.

\noindent{\bf Model Implementation.}
This flexibility allows BiFM to fine-tune large pretrained flow matching models, such as Stable Diffusion 3, for inversion-based image editing. During fine-tuning, we apply LoRA on model backbone, following settings from \citet{chadebec2025flash}. As shown in \Cref{fig:rethink_ddim}(a,b), for encoding time interval embedding input, we augment model backbone with an extra time embedding - we embed $t$ and $(t' - t)$ using standard MLP-based time embeddings, and then add them into a single interval embedding vector. This interval embedding is injected into the network in the same way as the original timestep embedding.

\noindent{\bf Inference with BiFM.}
At inference Time, BiFM performs inversion-based editing following \Cref{alg:bifm_edit}. For tradeoff between sampling quality and efficiency, BiFM supports one-step and multi-step sampling by decomposing large time intervals, as shown in \Cref{alg:code_sample}. 

In summary, BiFM unifies generation and inversion under a shared velocity field, enabling accurate few-step sampling and editing. Unlike standard diffusion models that rely on numerical ODE/SDE solvers or DDIM-based inversion, it avoids iterative solvers and DDIM approximation errors. Compared to MeanFlow, BiFM extends velocity supervision to both time directions, supporting joint training and fine-tuning for efficient and robust image editing.

\definecolor{codeblue}{rgb}{0.25,0.5,0.5}
\definecolor{codekw}{rgb}{0.85, 0.18, 0.50}

\definecolor{codesign}{RGB}{0, 0, 255}
\definecolor{codefunc}{rgb}{0.85, 0.18, 0.50}

\lstdefinelanguage{PythonFuncColor}{
  language=Python,
  keywordstyle=\color{blue}\bfseries,
  commentstyle=\color{codeblue},  
  stringstyle=\color{orange},
  showstringspaces=false,
  basicstyle=\ttfamily\small,
  literate=
    {*}{{\color{codesign}* }}{1}
    {-}{{\color{codesign}- }}{1}
    {+}{{\color{codesign}+ }}{1}
    {dataloader}{{\color{codefunc}dataloader}}{1}
    {timesampler}{{\color{codefunc}timesampler}}{1}
    {randn}{{\color{codefunc}randn}}{1}
    {randn_like}{{\color{codefunc}randn\_like}}{1}
    {jvp}{{\color{codefunc}jvp}}{1}
    {stopgrad}{{\color{codefunc}stopgrad}}{1}
    {metric}{{\color{codefunc}metric}}{1}
    {model}{{\color{codefunc}model}}{1}
}

\lstset{
  language=PythonFuncColor,
  backgroundcolor=\color{white},
  basicstyle=\fontsize{9pt}{9.9pt}\ttfamily\selectfont,
  columns=fullflexible,
  breaklines=true,
  captionpos=b,
}






\begin{algorithm}[h!]
\caption{BiFM: Inversion-Based Editing.}
\label{alg:bifm_edit}
\begin{lstlisting}[language=python]
# x_1: source image
# p_s: source prompt
# p_t: target prompt
u = model(x_1,1,0,p_s)                             # inversion
x_0 = x_1 + u
u_edit = model(x_0,0,1,p_t)                        # generation
x_1_edit = x_0 + u_edit
return x_1_edit
\end{lstlisting}
\end{algorithm}
\vspace{-12pt}
\definecolor{codeblue}{rgb}{0.25,0.5,0.5}
\definecolor{codekw}{rgb}{0.85, 0.18, 0.50}

\definecolor{codesign}{RGB}{0, 0, 255}
\definecolor{codefunc}{rgb}{0.85, 0.18, 0.50}

\lstdefinelanguage{PythonFuncColor}{
  language=Python,
  keywordstyle=\color{blue}\bfseries,
  commentstyle=\color{codeblue},  
  stringstyle=\color{orange},
  showstringspaces=false,
  basicstyle=\ttfamily\small,
  literate=
    {*}{{\color{codesign}* }}{1}
    {-}{{\color{codesign}- }}{1}
    {+}{{\color{codesign}+ }}{1}
    {dataloader}{{\color{codefunc}dataloader}}{1}
    {sample_t_r}{{\color{codefunc}sample\_t\_r}}{1}
    {randn}{{\color{codefunc}randn}}{1}
    {randn_like}{{\color{codefunc}randn\_like}}{1}
    {jvp}{{\color{codefunc}jvp}}{1}
    {stopgrad}{{\color{codefunc}stopgrad}}{1}
    {metric}{{\color{codefunc}metric}}{1}
    {linspace}{{\color{codefunc}linspace}}{1}
    {model}{{\color{codefunc}model}}{1}
}

\lstset{
  language=PythonFuncColor,
  backgroundcolor=\color{white},
  basicstyle=\fontsize{9pt}{9.9pt}\ttfamily\selectfont,
  columns=fullflexible,
  breaklines=true,
  captionpos=b,
}






        

\begin{algorithm}[h]
\caption{{BiFM}: Multi-Step Sampling.}
\label{alg:code_sample}
\begin{lstlisting}[language=python]
# N: number of sample steps
time_steps = linspace(0,1,N)
e = randn(x_shape)
x_0 = e
z = x_0
for i in range(N):
    t_s, t_e = time_steps[i:i+2]
    u = model(z,t_s,t_e)
    z = z + (t_e - t_s) * u
x_1 = z
return x_1
\end{lstlisting}
\end{algorithm}
\vspace{-8pt}

\section{Experiments}
\label{sec:experiments}

\begin{table*}[t!]
    \centering 
    \resizebox{0.75\textwidth}{!}{%
    \begin{tabular}{lllllllll}
        \toprule
        \multirow{2}{*}{\textbf{Settings}}  & \multirow{2}{*}{\textbf{Methods}} & \multirow{2}{*}{\textbf{NFE}} & \multicolumn{4}{c}{\textbf{Background Preservation}} & \multicolumn{2}{c}{\textbf{CLIP Semantics}} \\
        \cmidrule(lr){4-7} \cmidrule(lr){8-9}
        & & & {\bf LPIPS}$_{\times10^3}\downarrow$ & {\bf SSIM}$_\%\uparrow$ & {\bf PSNR}$\uparrow$  & {\bf MSE}$_{\times10^4}\downarrow$  & {\bf Whole}$\uparrow$ & {\bf Edited}$\uparrow$ \\ 
        
        \midrule
        \multirow{7}{*}{Multi-Step}
        & NT Inv~\citep{mokady2023null} (CVPR23)       & 50         & 60.67  & 84.11 & 27.03 & \underline{35.86}  & 23.61 & 21.64 \\
        & MasaCtrl~\citep{cao2023masactrl} (ICCV23)     & 50         & 106.62 & 79.67 & 22.17 & 86.97  & 23.96 & 21.16 \\ 
        & PnP Inv~\citep{ju2024pnp} (ICLR24)           & 50         & \underline{49.25}  & 84.86 & \underline{27.22} & {\bf 32.87}  & 25.83 & 22.39 \\
        & EditFT~\citep{xu2025editft} (CVPR25)           & 30 & 80.55  & {\bf 91.50} & 26.62 & 40.24 & {25.74} & 22.27 \\
        & FlowEdit~\citep{kulikov2025flowedit} (ICCV25)        & 28 & 112.19 & 83.08 & 21.96 & 94.99 & 25.25 & 22.58 \\        
        & DNAEdit~\citep{xie2025dnaedit} (NeurIPS25)           & 28 & 112.60 & 83.69 & 23.24 & 67.32 & {\bf 28.90} & \underline{23.66} \\        
        & {\bf BiFM} (ours)                    & 50         & \textbf{47.01}  & \underline{87.50} & \textbf{29.89} & 42.66  & \underline{27.42} & \textbf{24.66} \\
            
        \midrule
        \multirow{6}{*}{Few-Step}  
        & DDIM~\citep{song21ddim} (ICLR21)          & 4         & 177.96 & 66.86 & 18.59 & 184.69 & 23.62 & 21.20 \\
        & TurboEdit~\citep{wu2024turboedit} (ECCV24) & 4    & {76.95}  & {84.63} & {25.51} & 58.48  & {25.49} & {21.82} \\
        & FireFlow~\citep{deng2024fireflow} (ICML25)          & 18 & -     & 82.49 & 23.03 & -   & 26.02 & 22.81 \\        
        & InstantEdit~\citep{gong2025instantedit} (ICCV25)    & 4  & {\bf 44.39} & \underline{86.44} & \underline{27.96} & {\bf 34.94} & \underline{26.28} & \underline{22.82} \\
        & ReNoise~\citep{garibi2024renoise} (NeurIPS25)  & 4     & 136.52 & 76.51 & 22.59 & \underline{53.49}  & 24.78 & 21.37 \\       
        & {\bf BiFM} (ours)                    & 4     & \underline{67.25}  & \textbf{87.29} & \textbf{28.92} & {54.92}  & \textbf{26.77} & \textbf{23.58} \\
        
        \midrule
        \multirow{2}{*}{One-Step} 
        & SwiftEdit~\citep{nguyen2025swiftedit} (CVPR25)  & 1  & \textbf{91.04}  & 81.05 & 23.33 & 66.58  & 25.16 & 21.25 \\
        & {\bf BiFM} (ours)                      & 1  & 92.30  & \textbf{85.88} & \textbf{28.46} & \textbf{58.22}  & \textbf{26.09} & \textbf{22.42} \\
        
        \bottomrule
    
    \end{tabular}
    }
    {
    \captionsetup{width=.95\linewidth}
    \caption{\small {\bf PIE-Bench Image Editing Performance.} We compare with baselines under different sampling budgets with evaluation metrics from PIE-Bench~\citep{ju2024pnp}. More baselines can be found in Appendix.}
    \label{tab:edit_piebench}
    }
    
\end{table*}

We evaluate BiFM as a unified framework for image generation and inversion-based image editing.
Our experiments aim to show that: (i) BiFM achieves accurate inversion and reconstruction, (ii) BiFM can be effectively applied to pretrained flow matching models for few-step prompt-guided image editing, and (iii)
BiFM enhances image generation sample quality across diverse  experimental settings. 
We also conduct ablation studies on image generation to analyze key design choices and their impact.

\subsection{Experiment Setup}

\noindent{\bf Image Editing.}
We fine-tune Stable Diffusion 3~\citep{esser2024sd3}, a state-of-the-art flow matching model, with BiFM and evaluate it on the PIE-Bench~\citep{ju2024pnp} benchmark.
For image inversion and reconstruction experiment, we use source images and their corresponding prompts as both input and target.
For prompt-guided image editing, we compare BiFM against both training-free inversion methods and few-step editing methods that require fine-tuning or distillation.
Evaluation metrics include MSE, PSNR, SSIM, and LPIPS. For semantic alignment, we use CLIP score.

\noindent{\bf Image Generation.}
For text-to-image generation on MSCOCO-256~\citep{lin2014coco}, we use an MMDiT backbone from REPA~\citep{yu2025repa} for training from scratch. To validate our design choices, we also assess class-conditional generation on ImageNet-256~\citep{deng2009imagenet} (fine-tuning SiT-XL/2) and small-resolution dataset (CIFAR-10~\citep{krizhevsky2009cifar}) trained from scratch with a U-Net backbone.
We evaluate image generation performance by FID, IS, Precision, and Recall.

\begin{table}[h!]  
    \centering 
    \resizebox{0.45\textwidth}{!}{%
    \begin{tabular}{l l l l l}
        \toprule
        \textbf{Method} & {\bf MSE$_{\times 10^4}^\downarrow$} & {\bf LPIPS$_{\times 10^3}^\downarrow$} & {\bf SSIM$_\%^\uparrow$} & {\bf PSNR$\uparrow$} \\
        
        \midrule
        DDIM~\citep{song21ddim} (ICLR21)          & 224.43 & 210.84 & 70.96 & 17.76 \\
        NT Inv~\citep{mokady2023null} (CVPR23)    & 97.64  & 102.25 & 82.33 & 27.82 \\
        PnP Inv~\citep{ju2024pnp} (ICLR24)        & 105.66 & \underline{95.95}  & \underline{87.20} & \underline{28.79} \\
        RF-Solver~\citep{wang2025rfedit} (ICML25) & \underline{94.80}  & 106.15 & 86.36 & 28.26 \\

        \textbf{BiFM} (ours)             & {\bf 87.72}  & {\bf 89.49}  & {\bf 88.03} & {\bf 30.32} \\
        
        \arrayrulecolor{black}\bottomrule        
    \end{tabular}        
    }
    \caption{\small {\bf Image Reconstruction Performance}. We use 50 inversion steps to generate results for all methods. BiFM’s learned inversion process greatly reduces reconstruction error.}
    \label{tab:inv_recon}
    \vspace{-6pt}
\end{table}

\subsection{Inversion and Reconstruction}

To demonstrate that BiFM learns accurate inversion of images, we evaluate image reconstruction results from different inversion methods as well as BiFM itself. We use background preservation metrics from PIE-Bench computed between source images and reconstructed images. 

\begin{figure}[ht]
    \centering
    \setlength{\abovecaptionskip}{6pt}
    \setlength{\tabcolsep}{1.5pt}
    \begin{tikzpicture}
    \node[inner sep=0pt] (table) {%
    \begin{tabular}{c c c c}

        Original & PnP Inv & RF-Edit & {\bf BiFM (ours)} \\

        \includegraphics[width=0.11\textwidth,height=0.11\textwidth]{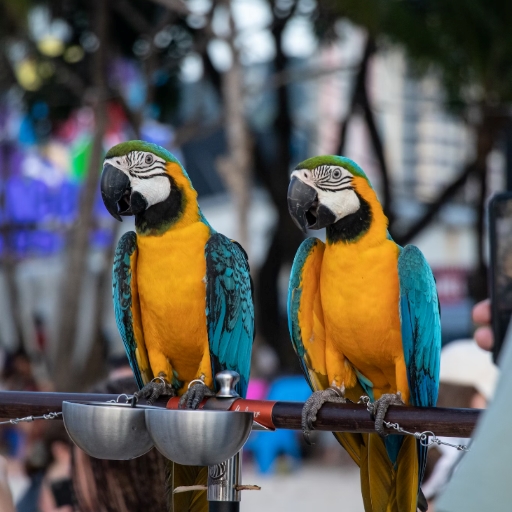} &
        \includegraphics[width=0.11\textwidth,height=0.11\textwidth]{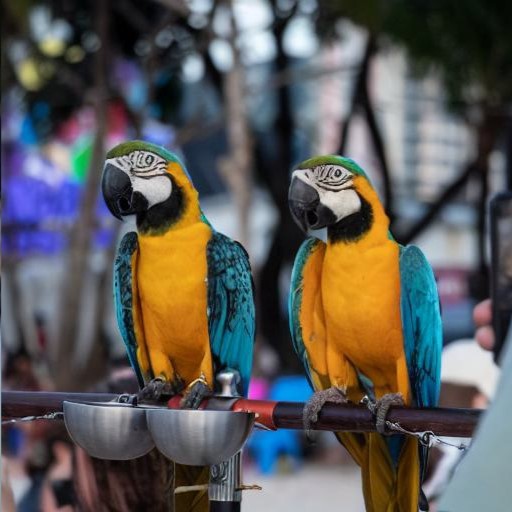} &
        \includegraphics[width=0.11\textwidth,height=0.11\textwidth]{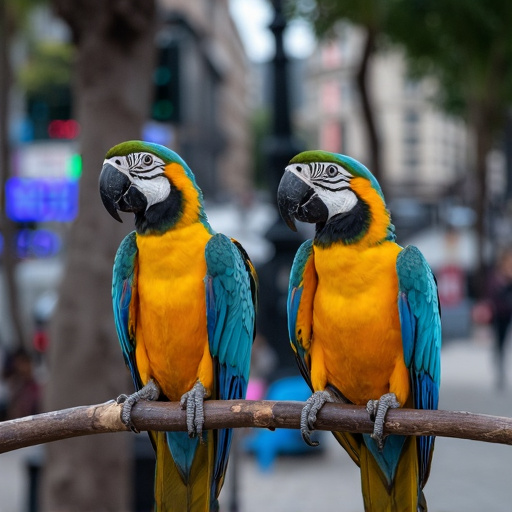} &
        \includegraphics[width=0.11\textwidth,height=0.11\textwidth]{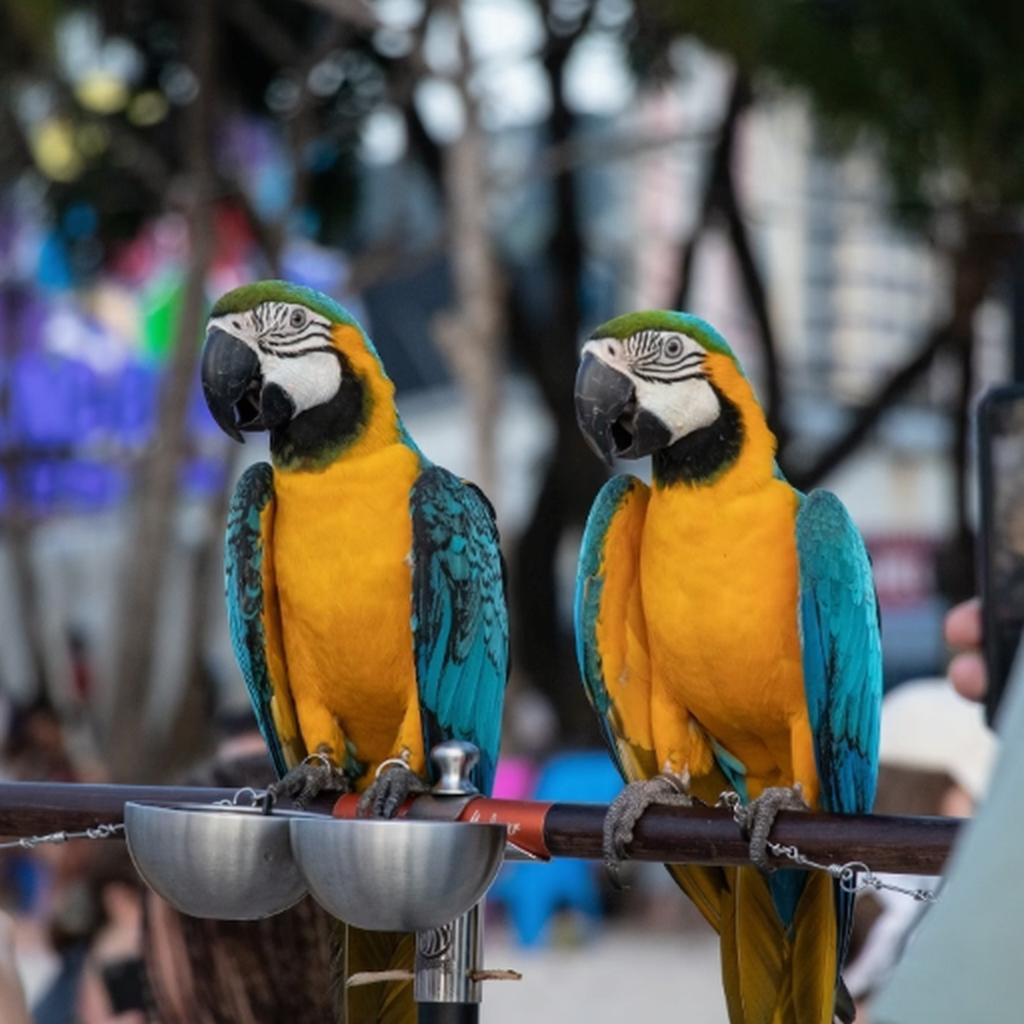} \\ 

        \includegraphics[width=0.11\textwidth,height=0.11\textwidth]{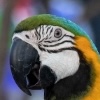} &
        \includegraphics[width=0.11\textwidth,height=0.11\textwidth]{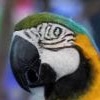} &
        \includegraphics[width=0.11\textwidth,height=0.11\textwidth]{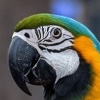} &
        \includegraphics[width=0.11\textwidth,height=0.11\textwidth]{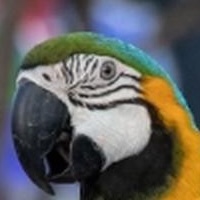} \\ 

        \includegraphics[width=0.11\textwidth,height=0.11\textwidth]{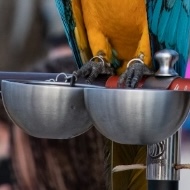} &
        \includegraphics[width=0.11\textwidth,height=0.11\textwidth]{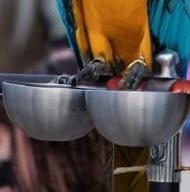} &
        \includegraphics[width=0.11\textwidth,height=0.11\textwidth]{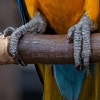} &
        \includegraphics[width=0.11\textwidth,height=0.11\textwidth]{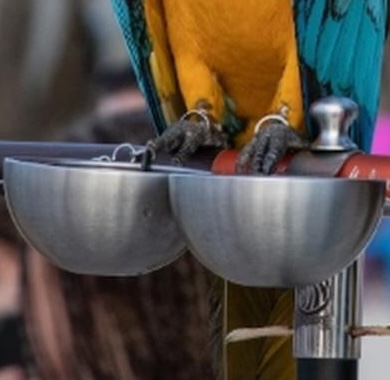}
        
    \end{tabular}
    };

    \draw[rered, ultra thick] ([xshift=-0pt,yshift=2pt]table.south east) rectangle ([xshift=-58pt,yshift=-11pt]table.north east);
    \end{tikzpicture}
    \caption{{\bf Inversion and Reconstruction Quality.} From left to right: original input image, PnP Inversion~\cite{ju2024pnp}, RF-Edit~\cite{wang2025rfedit}, and \textbf{BiFM (ours)}. 
    BiFM faithfully reconstructs image details, while RF-Edit exhibits semantic shift and PnP Inv fails to recover fine details in the source image.}
    \label{fig:visual_inv_recon}
    \vspace{-14pt}
\end{figure}

As shown in \Cref{tab:inv_recon}, BiFM achieves the best performance on all evaluation metrics, outperforming baselines by a clear margin. In \Cref{fig:visual_inv_recon}, compared to PnP Inversion and RF-Edit, BiFM preserves global scene layout while recovering sharper local details (such as eyes and object geometries), indicating consistent bidirectional flow learning.

\begin{figure*}[ht]
    \centering
    \setlength{\abovecaptionskip}{6pt}
    \setlength{\tabcolsep}{1.5pt}
    \begin{tikzpicture}
    \node[inner sep=0pt] (table) {%
    \begin{tabular}{c c c c c c c c}

        Prompt & Original & PnP Inv & RF-Edit & FlowEdit & ReNoise & SwiftEdit & {\bf BiFM (ours)} \\

        \raisebox{0.04\textwidth}{\rotatebox[origin=t]{0}{\scalebox{0.8}{\begin{tabular}{c@{}c@{}c@{}} \emph{``tulip$\rightarrow$lion"} \end{tabular}}}} & 
        \includegraphics[width=0.12\textwidth,height=0.12\textwidth]{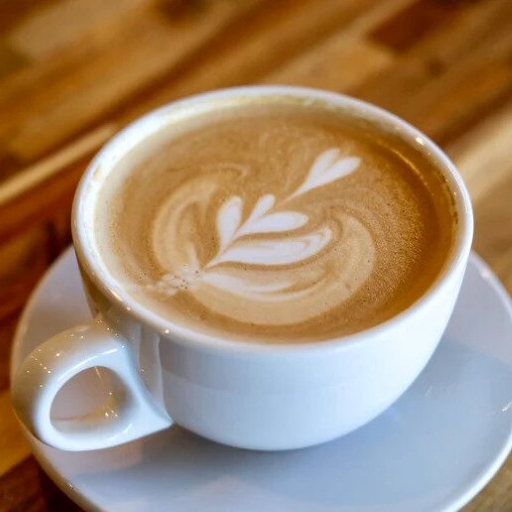} &
        \includegraphics[width=0.12\textwidth,height=0.12\textwidth]{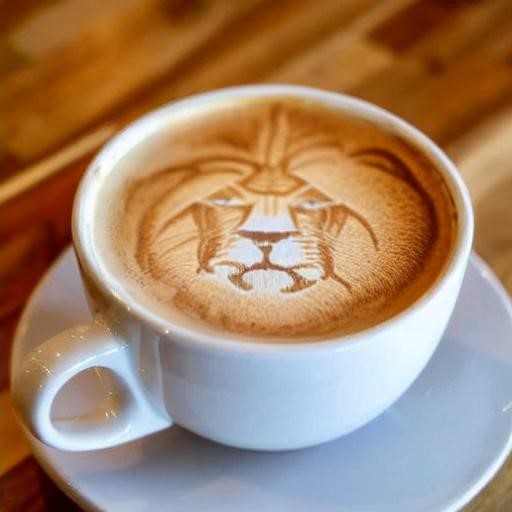} &
        \includegraphics[width=0.12\textwidth,height=0.12\textwidth]{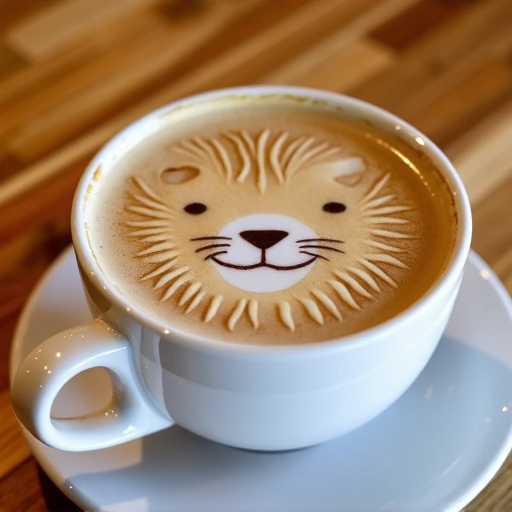} &
        \includegraphics[width=0.12\textwidth,height=0.12\textwidth]{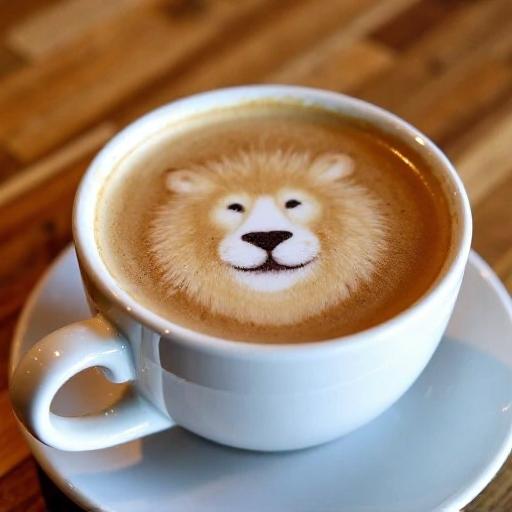} &
        \includegraphics[width=0.12\textwidth,height=0.12\textwidth]{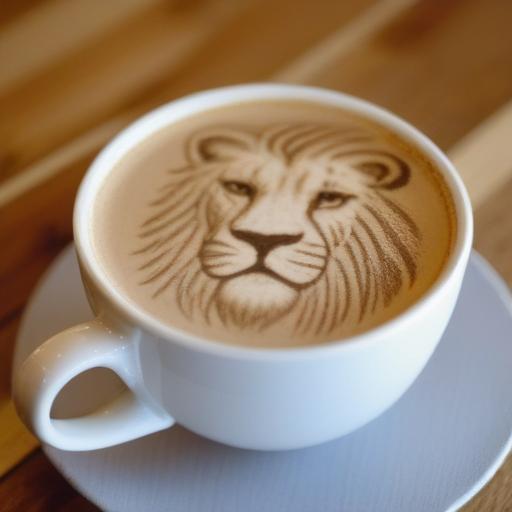} &
        \includegraphics[width=0.12\textwidth,height=0.12\textwidth]{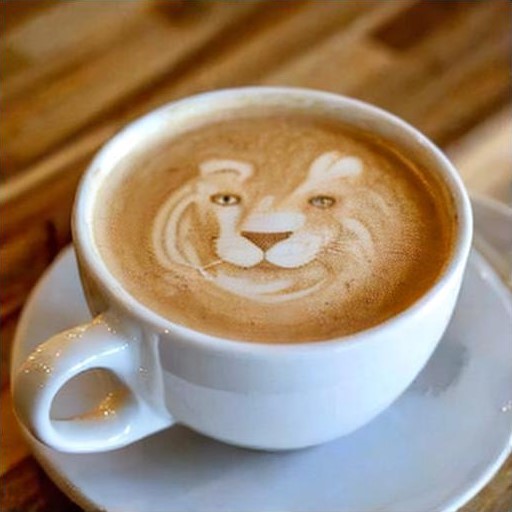} &
        \includegraphics[width=0.12\textwidth,height=0.12\textwidth]{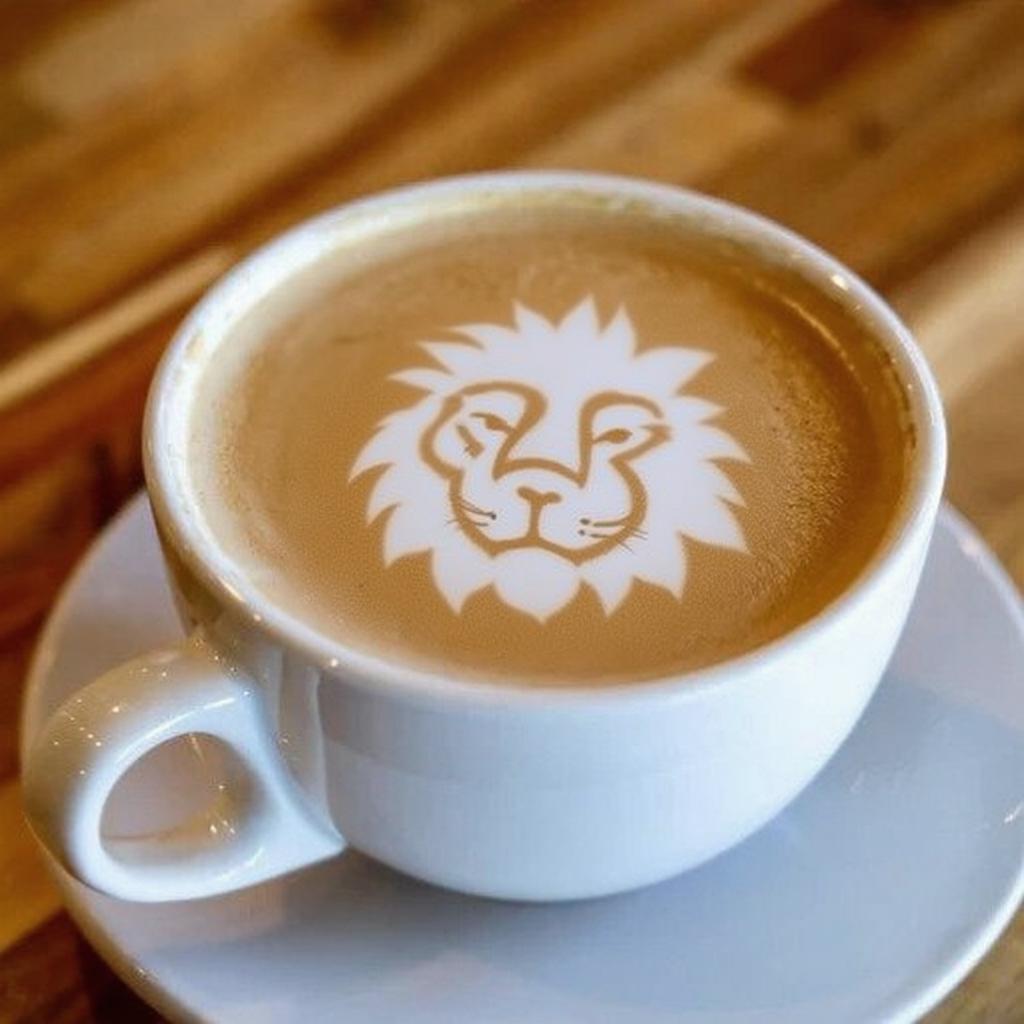} \\ 

        \raisebox{0.03\textwidth}{\rotatebox[origin=t]{0}{\scalebox{0.8}{\begin{tabular}{c@{}c@{}c@{}} \emph{``colorful$\rightarrow$red"} \end{tabular}}}} &
        \includegraphics[width=0.12\textwidth,height=0.12\textwidth]{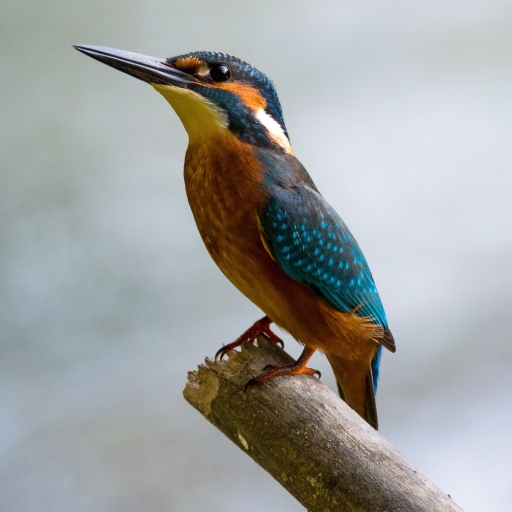} &
        \includegraphics[width=0.12\textwidth,height=0.12\textwidth]{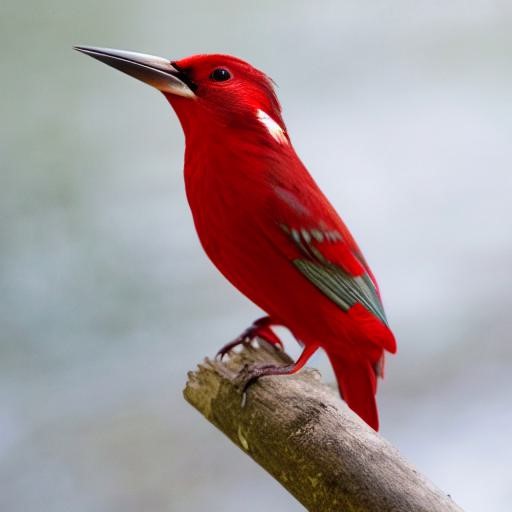} &
        \includegraphics[width=0.12\textwidth,height=0.12\textwidth]{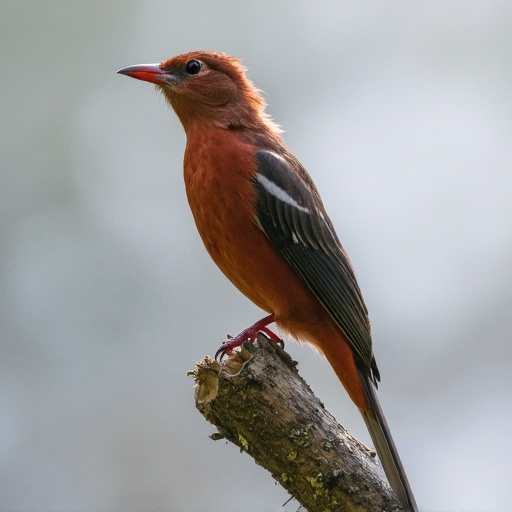} &
        \includegraphics[width=0.12\textwidth,height=0.12\textwidth]{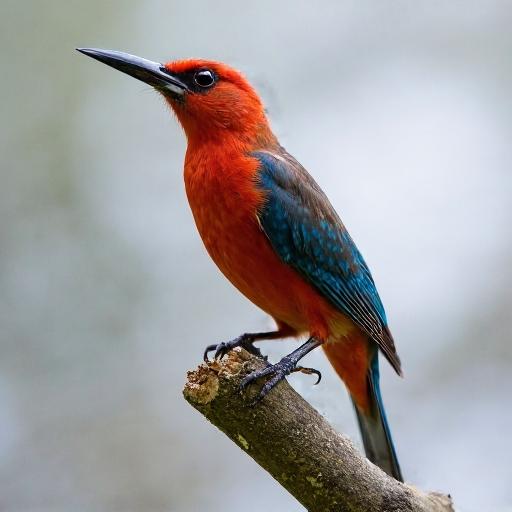} &
        \includegraphics[width=0.12\textwidth,height=0.12\textwidth]{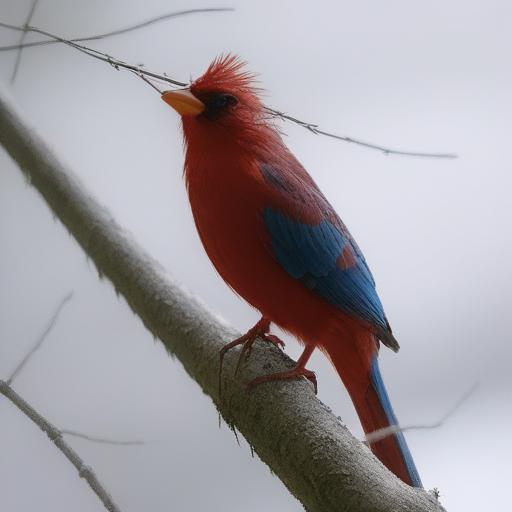} &
        \includegraphics[width=0.12\textwidth,height=0.12\textwidth]{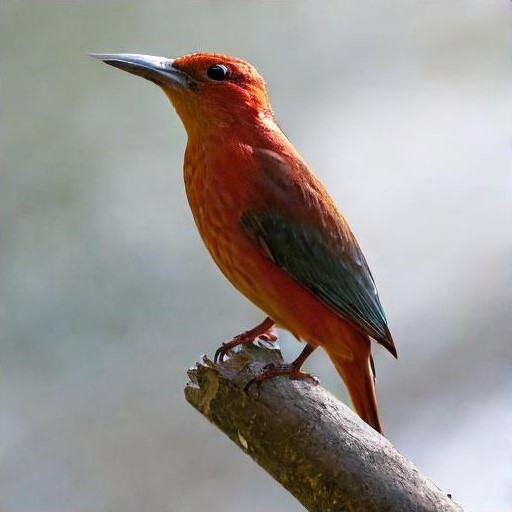} &
        \includegraphics[width=0.12\textwidth,height=0.12\textwidth]{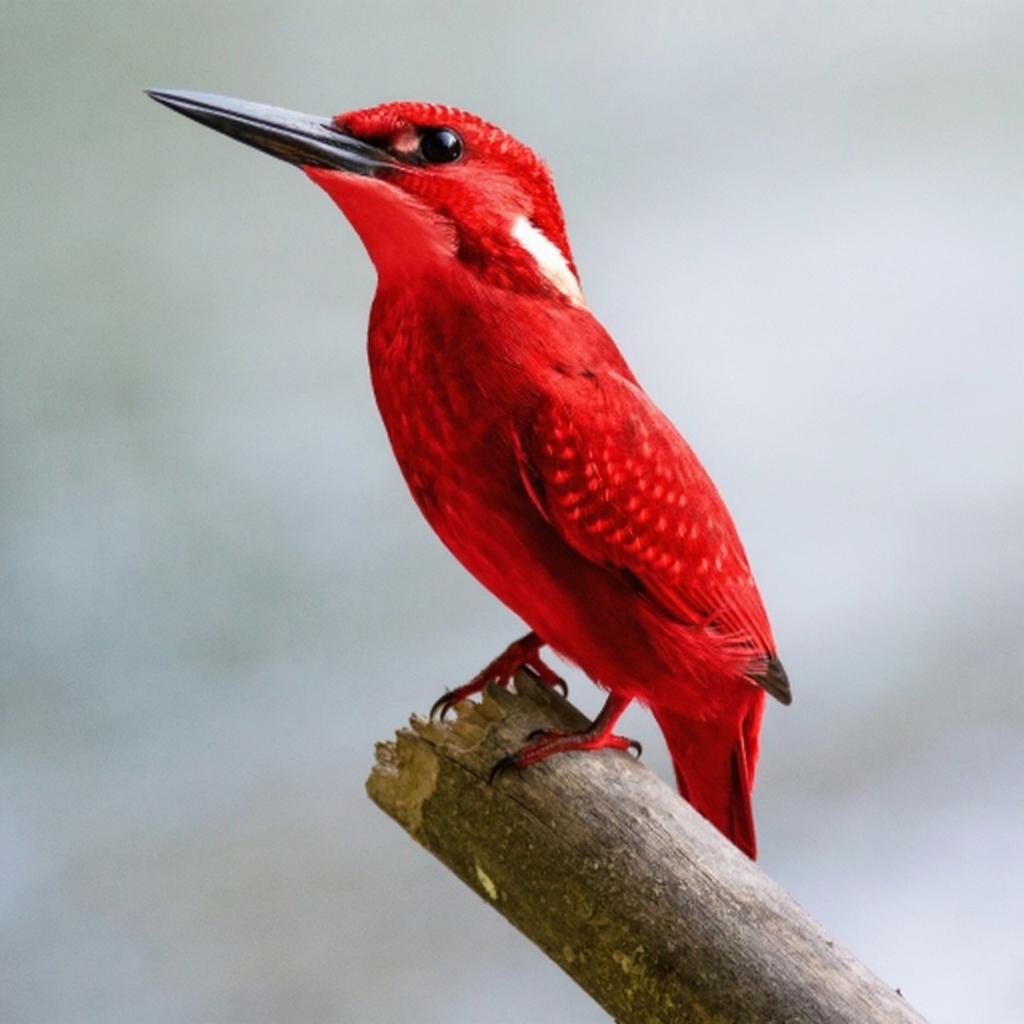} \\ 

        \raisebox{0.03\textwidth}{\rotatebox[origin=t]{0}{\scalebox{0.8}{\begin{tabular}{c@{}} \emph{``fruits$\rightarrow$pizza"} \end{tabular}}}} &
        \includegraphics[width=0.12\textwidth,height=0.12\textwidth]{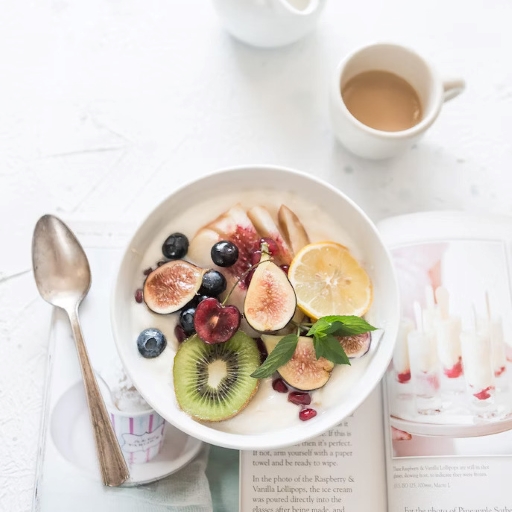} &
        \includegraphics[width=0.12\textwidth,height=0.12\textwidth]{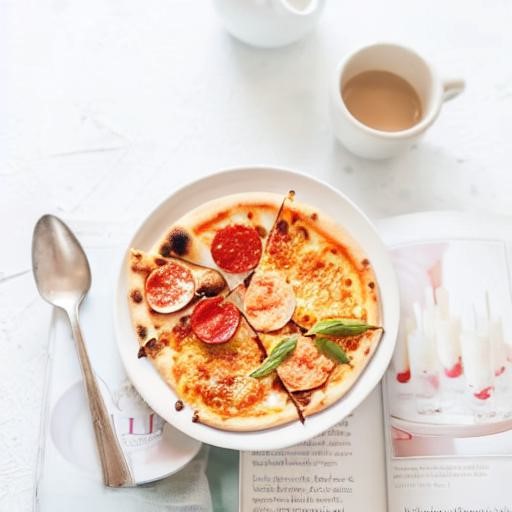} &
        \includegraphics[width=0.12\textwidth,height=0.12\textwidth]{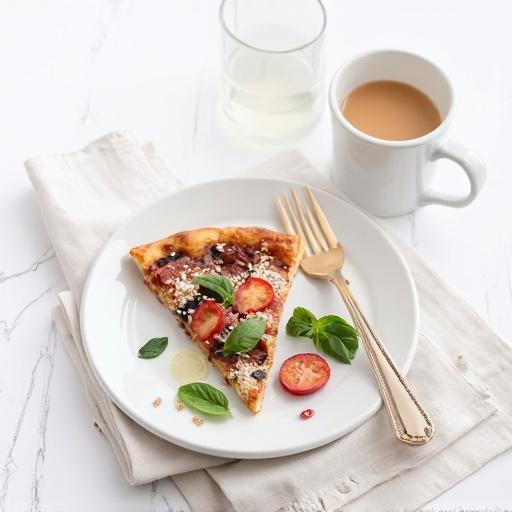} &
        \includegraphics[width=0.12\textwidth,height=0.12\textwidth]{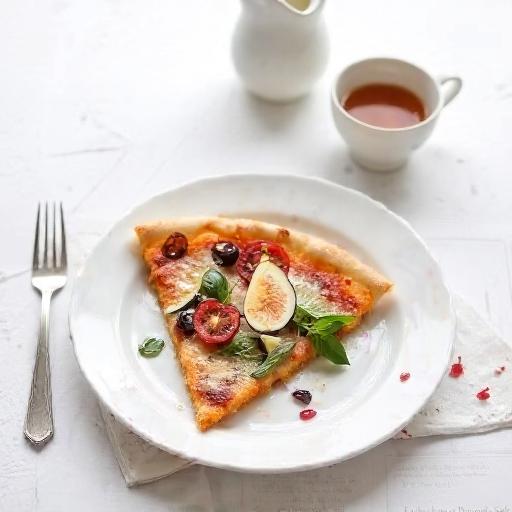} &
        \includegraphics[width=0.12\textwidth,height=0.12\textwidth]{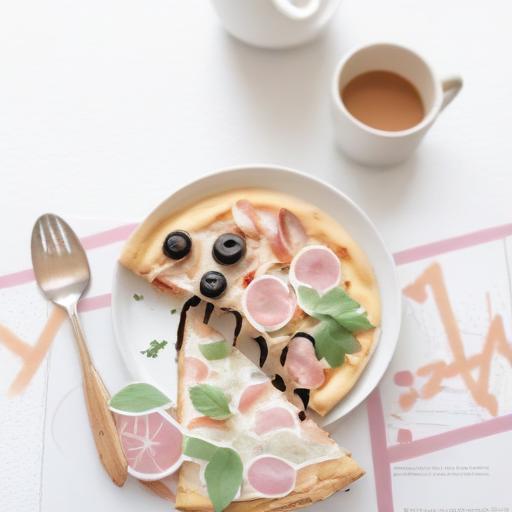} &
        \includegraphics[width=0.12\textwidth,height=0.12\textwidth]{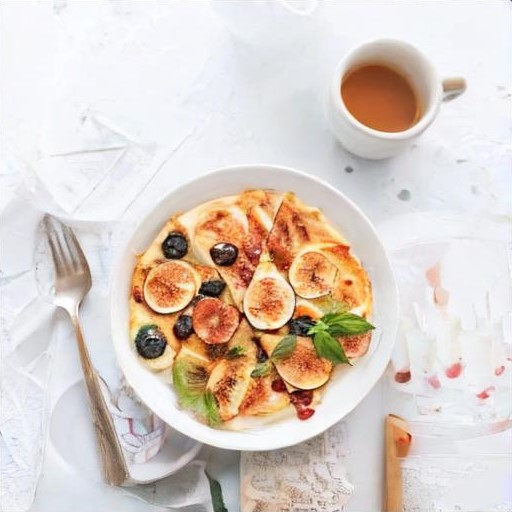} &
        \includegraphics[width=0.12\textwidth,height=0.12\textwidth]{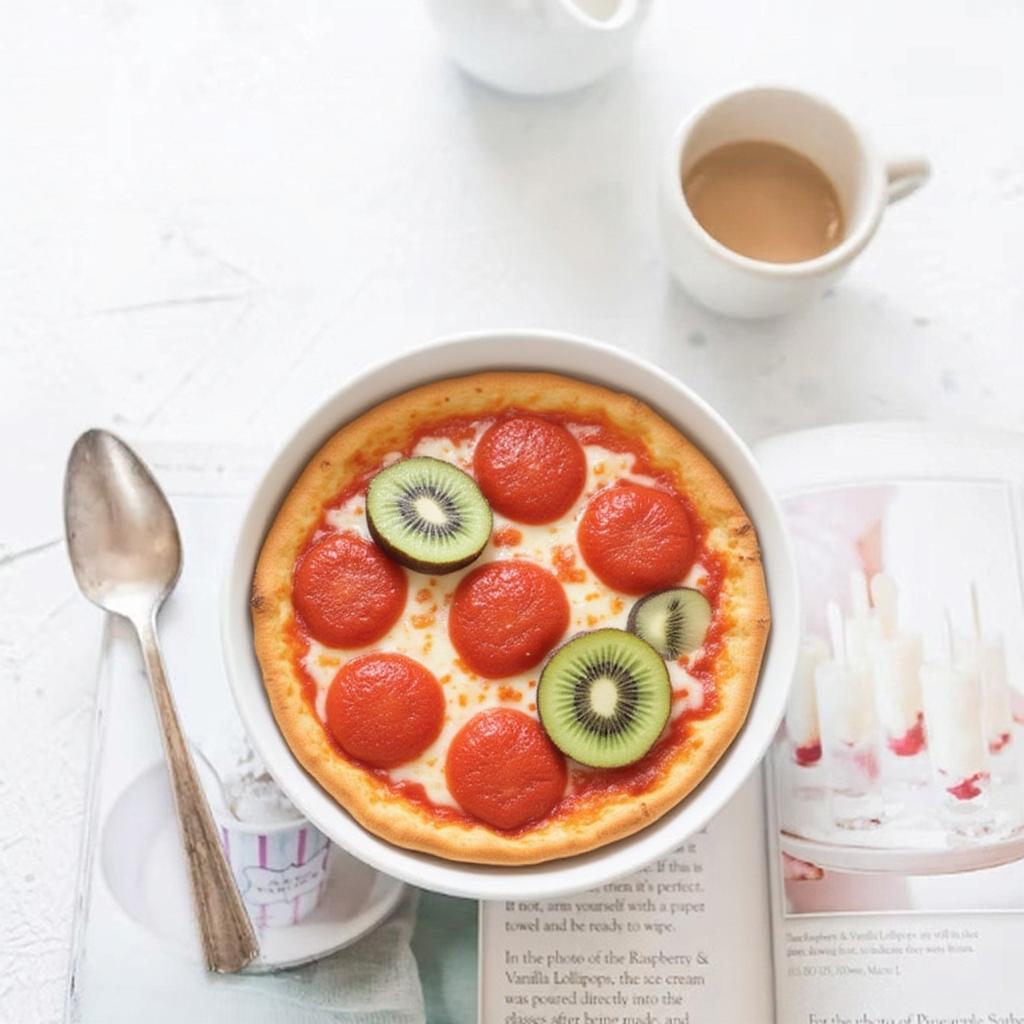} \\
        
        \raisebox{0.04\textwidth}{\rotatebox[origin=t]{0}{\scalebox{0.8}{\begin{tabular}{c@{}c@{}c@{}} \emph{``torch$\rightarrow$flower"} \end{tabular}}}} &
        \includegraphics[width=0.12\textwidth,height=0.12\textwidth]{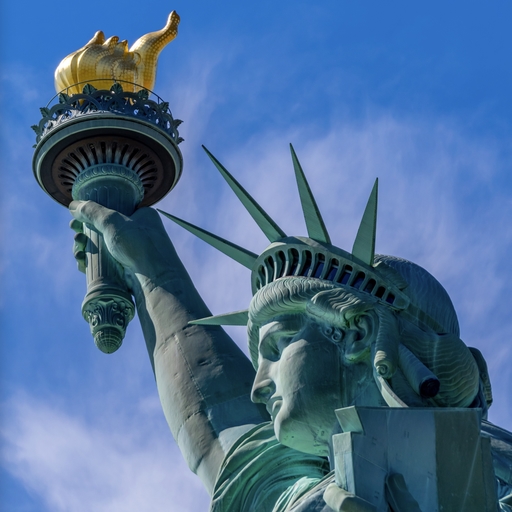} &
        \includegraphics[width=0.12\textwidth,height=0.12\textwidth]{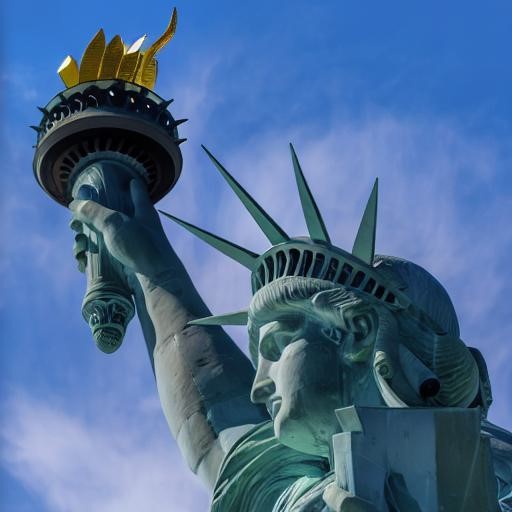} &
        \includegraphics[width=0.12\textwidth,height=0.12\textwidth]{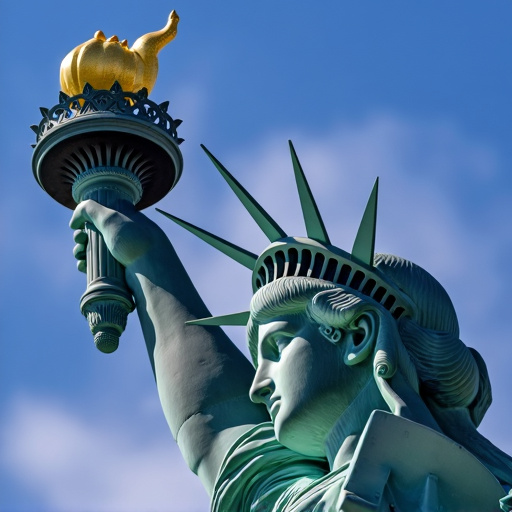} &
        \includegraphics[width=0.12\textwidth,height=0.12\textwidth]{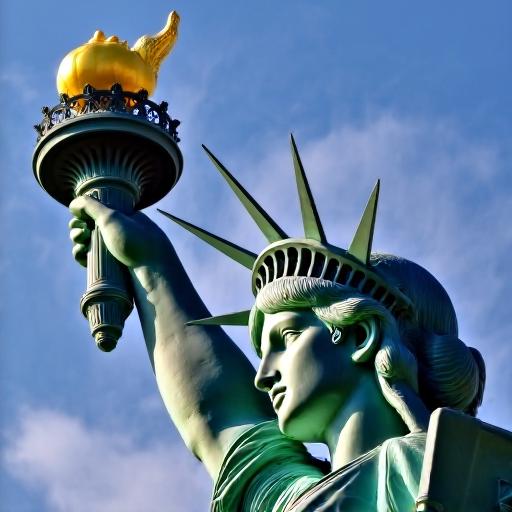} &
        \includegraphics[width=0.12\textwidth,height=0.12\textwidth]{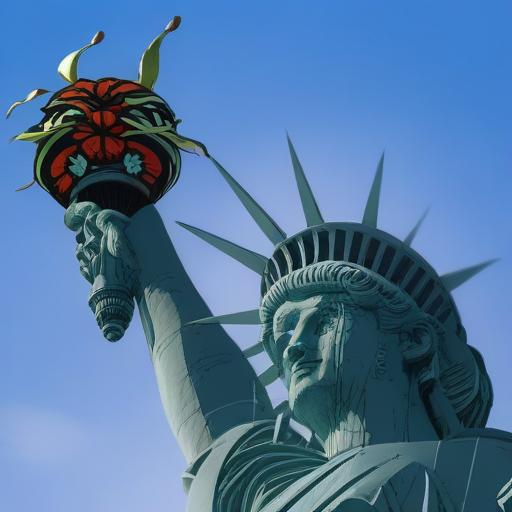} &
        \includegraphics[width=0.12\textwidth,height=0.12\textwidth]{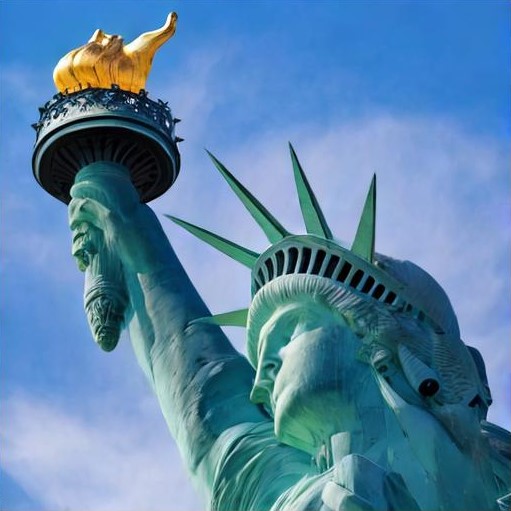} &
        \includegraphics[width=0.12\textwidth,height=0.12\textwidth]{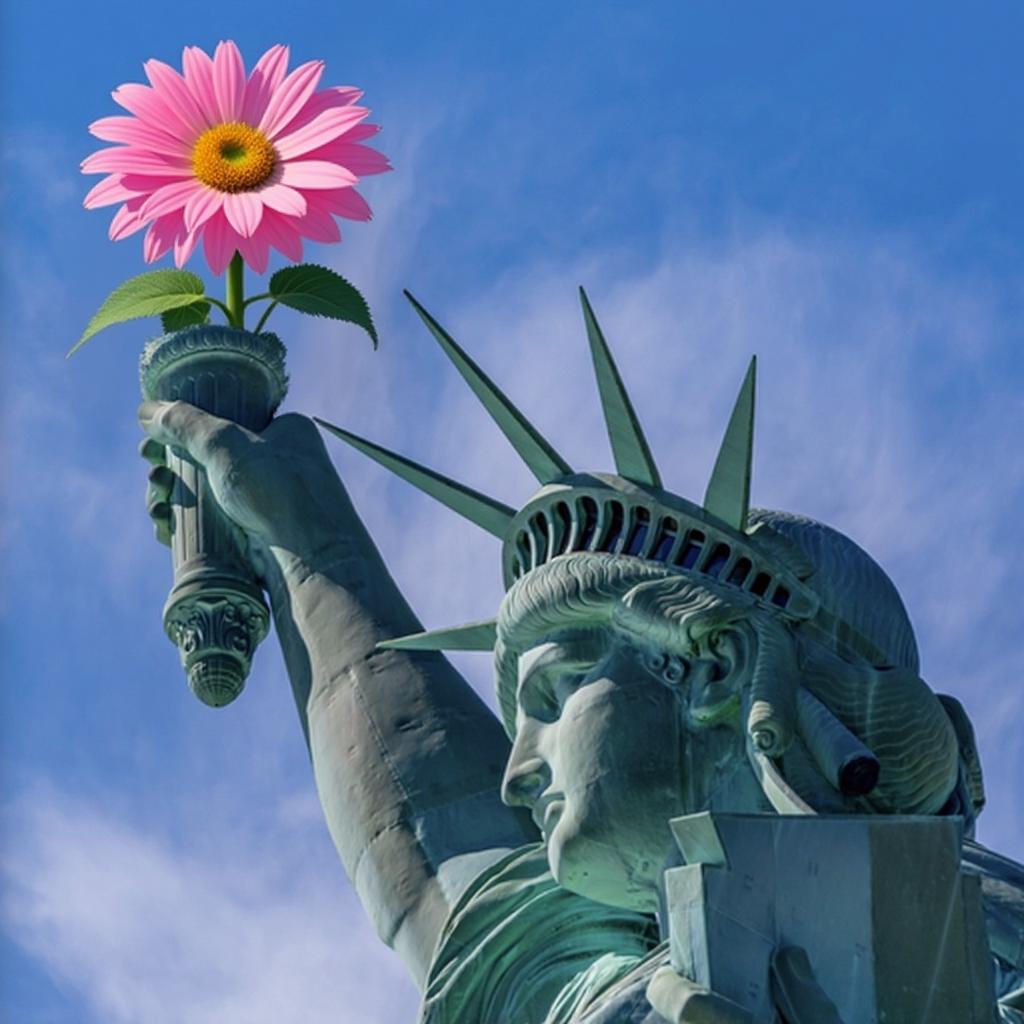} \\

        \raisebox{0.04\textwidth}{\rotatebox[origin=t]{0}{\scalebox{0.8}{\begin{tabular}{c@{}c@{}c@{}} \emph{``sea$\rightarrow$forest"} \end{tabular}}}} &
        \includegraphics[width=0.12\textwidth,height=0.12\textwidth]{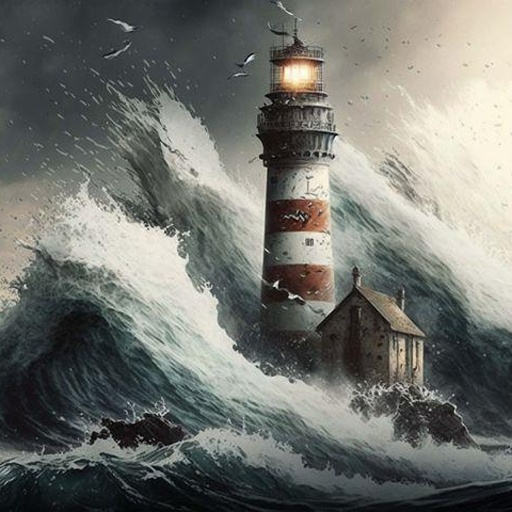} &
        \includegraphics[width=0.12\textwidth,height=0.12\textwidth]{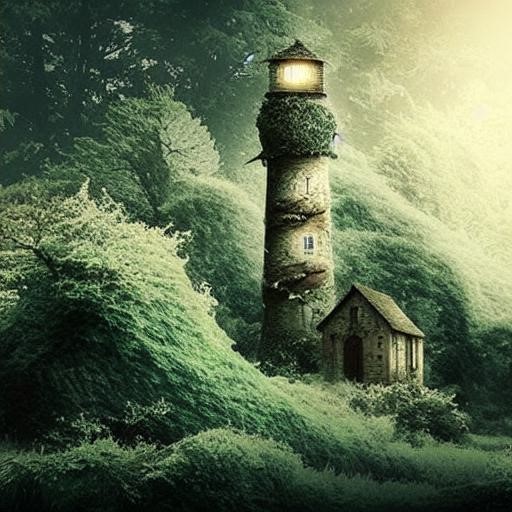} &
        \includegraphics[width=0.12\textwidth,height=0.12\textwidth]{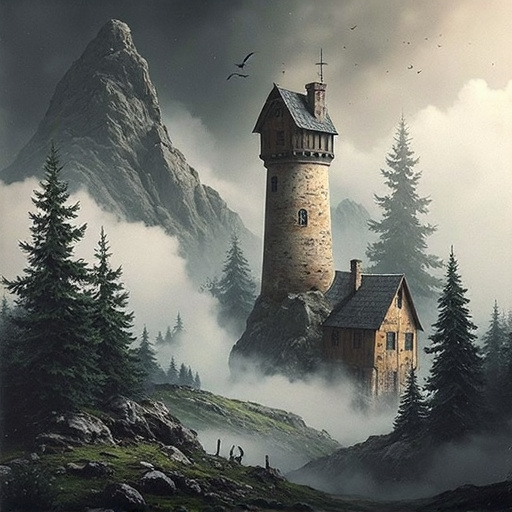} &
        \includegraphics[width=0.12\textwidth,height=0.12\textwidth]{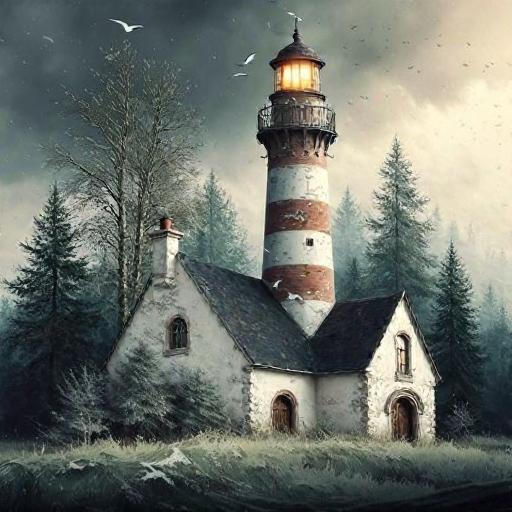} &
        \includegraphics[width=0.12\textwidth,height=0.12\textwidth]{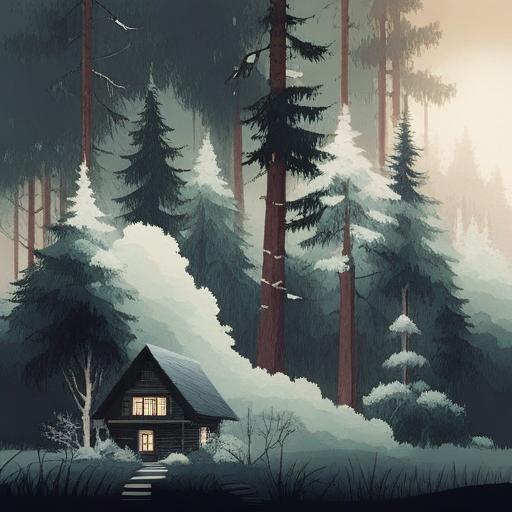} &
        \includegraphics[width=0.12\textwidth,height=0.12\textwidth]{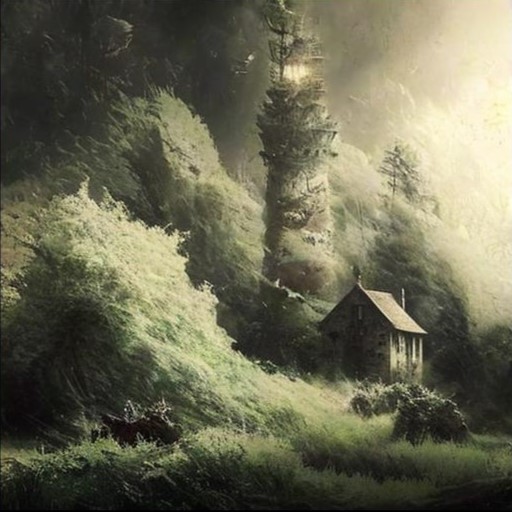} &
        \includegraphics[width=0.12\textwidth,height=0.12\textwidth]{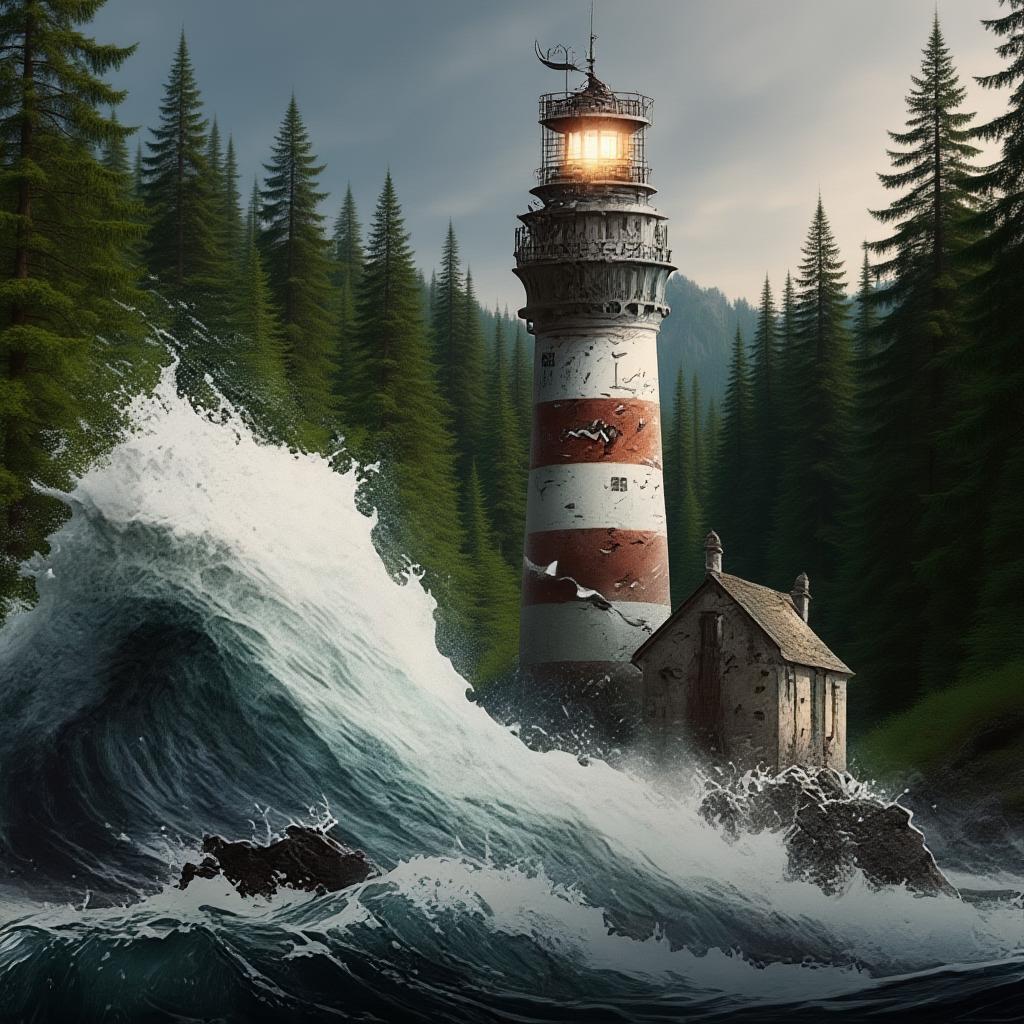} \\

    \end{tabular}
    };

    \draw[rered, ultra thick] ([xshift=-0pt,yshift=2pt]table.south east) rectangle ([xshift=-63pt,yshift=-11pt]table.north east);
    \end{tikzpicture}
    \caption{\small {\bf Image Editing Visualization.} Given a source image, a source prompt and a target prompt (left illustrates difference between source prompt and target prompt), BiFM generates edits which follow more faithfully the intended concept while better preserving the original layout and fine details than other baselines. For example, BiFM engraves a clear {\it lion} pattern on the latte art without distorting the background, swaps the {\it Statue of Liberty’s} torch for a flower without geometry distortion, and maintains the {\it lighthouse} structure.} 
    \label{fig:edit_visual}
\end{figure*}

\subsection{Prompt-Based Image Editing}

\noindent{\bf Model Configuration.}
We fine-tune a pretrained Stable Diffusion 3 model using BiFM. We use the  MagicBrush~\citep{zhang2023magicbrush} training dataset for fine-tuning the image editing models.
For configuring LoRA, we follow the LoRA hyperparameter settings from Flash Diffusion~\citep{chadebec2025flash}.
Model weights are augmented with an extra time embedding for time-interval representation. The architecture of this additional embedding is identical to the model's original time embedding and is zero-initialized for warm-up. See more model training details in the  Appendix.

\noindent{\bf Baseline Comparison.}
We compare BiFM with existing image editing methods, including training-free methods and few-step editing methods, under three sampling budgets: multi-step, few-step, and one-step. \Cref{tab:edit_piebench} summarizes the results on PIE-Bench across background preservation and CLIP-based semantic metrics.

\noindent{\bf Multi-Step.} In the multi-step setting (50 NFEs), BiFM achieves the best overall balance between reconstruction fidelity and semantic alignment. Compared with training-free baselines such as MasaCtrl~\citep{cao2023masactrl} and PnP Inv~\citep{ju2024pnp}, BiFM improves both background preservation metrics and CLIP semantics, indicating that it better preserves background content while producing more faithful edits.

\noindent{\bf Few-Step \& One-Step.} Under the 4-step regime, BiFM attains LPIPS $67.25$, SSIM $87.29$, and PSNR $28.92$, outperforming both training-free inversion~\citep{song21ddim,garibi2024renoise} and auxiliary network methods~\citep{wu2024turboedit}, while remaining competitive on MSE. This supports our claim that explicitly learning forward/backward average velocities within a single network stabilizes few-step inversion-editing. Compared to SwiftEdit~\citep{nguyen2025swiftedit}, BiFM trades a slightly higher LPIPS ($92.30 ~vs.~91.04$) for materially better SSIM, PSNR, MSE, and CLIP semantics. This suggests BiFM favors structural/semantic preservation in the extreme one-step regime.

\begin{table*}[t!]
    \centering\small
    
    \begin{subtable}[t]{0.23\linewidth}
        \centering
        \begin{tabular}{p{0.55\linewidth} p{0.20\linewidth}}
            \toprule
            {\bf Time Input}        & {\bf FID$\downarrow$} \\
            \midrule
            $(t,t',\text{direc.})$  & 69.01                 \\
            $t'-t$                  & 60.86                 \\
            $(t, t')$               & 59.37                 \\
            \arrayrulecolor{black!40}\midrule
            $(t, t'-t)$             & \textbf{55.22}        \\
            \arrayrulecolor{black}\bottomrule
        \end{tabular}
        \caption{\footnotesize\textbf{Timestep embedding.} The model is conditioned on specific time parameterization designs.}
        \label{subtab:time_embed}
    \end{subtable}
    \hfill
    \begin{subtable}[t]{0.23\linewidth}
        \centering
        \begin{tabular}{p{0.60\linewidth} p{0.15\linewidth}}
            \toprule
            {\bf Time Sampler}  & {\bf FID$\downarrow$} \\
            \midrule
            uniform(0, 1)       & 63.25                 \\
            lognorm(--0.4, 1.2) & 59.16                 \\
            lognorm(--0.2, 1.0) & 58.57                 \\
            \arrayrulecolor{black!40}\midrule
            lognorm(--0.4, 1.0) & {\bf 55.22}           \\
            \arrayrulecolor{black}\bottomrule
        \end{tabular}
        \caption{\footnotesize\textbf{Time sampler.} $t$ and $t'$ are sampled from the specific distributions during training.}
        \label{subtab:time_sampler}
    \end{subtable}
    \hfill
    \begin{subtable}[t]{0.23\linewidth}
        \centering
        \begin{tabular}{p{0.60\linewidth} p{0.15\linewidth}}
            \toprule
            {\bf Weighting Func.} & {\bf FID$\downarrow$} \\
            \midrule
            linear                & 67.37                 \\
            sin                   & 62.45                 \\
            log                   & 57.93                 \\
            \arrayrulecolor{black!40}\midrule
            warm up               & {\bf 55.22}           \\
            \arrayrulecolor{black}\bottomrule
        \end{tabular}
        \caption{\footnotesize\textbf{Weighting function}. A warm-up schedule on the consistency term yields the best performance.}
        \label{subtab:weighting}
    \end{subtable}
    \hfill
    \begin{subtable}[t]{0.23\linewidth}
        \centering
        \begin{tabular}{p{0.55\linewidth} p{0.20\linewidth}}
            \toprule
            {\bf Loss Norm} & {\bf FID$\downarrow$} \\
            \midrule
            0.0             & 72.84                 \\
            2.0             & 59.66                 \\
            0.5             & 58.74                 \\
            \arrayrulecolor{black!40}\midrule
            1.0             & {\bf 55.22}           \\
            \arrayrulecolor{black}\bottomrule
        \end{tabular}
        \caption{\footnotesize\textbf{Loss norm metrics.} $p{=}0$ is squared L2 loss. $p{=}0.5$ is Pseudo-Huber loss.}
        \label{subtab:norm}
    \end{subtable}
    \vspace{-0.5em}
    \caption{\textbf{Ablation Study on 1-NFE Image Generation.} FID computed from 50K samples is reported. Defaults are shown in bottom row.}
    \label{tab:ablation}
\end{table*}

\noindent{\bf Editing Visualization.}
As shown in \Cref{fig:edit_visual}, BiFM produces edits that better satisfy the target prompts while more faithfully preserving object structure and background details than baselines.

\subsection{Image Generation}

We conduct image generation experiments on both small and large resolution datasets. We explore two training settings for BiFM: training from scratch and fine-tuning.

\noindent{\bf Text-to-Image Generation:} For text-to-image generation on MSCOCO-256, we train a vanilla MMDiT following the~\citep{yu2025repa} configuration and evaluate FID on 40k samples (equal to the validation set size) using REPA configurations without representation alignment.
\Cref{tab:t2i} shows evaluation results on FID. With MMDiT, BiFM reduces FID from $6.05$ (vanilla) and $4.73$ (REPA) to $4.57$, validating that our bidirectional average-velocity training complements representation-alignment style improvements.

\begin{table}[h!]
    \centering\small
    \begin{tabular}{l l l}
    
    \toprule
    {\bf Method} & {\bf Model} & {\bf FID$\downarrow$} \\
    
    \midrule
    U-Net~\citep{ronneberger2015unet}    & Diffusion     & 7.32 \\
    U-ViT-S/2~\citep{bao2022uvit}        & Diffusion     & 5.95 \\
    U-ViT-S/2 (Deep)~\citep{bao2022uvit} & Diffusion     & 5.48 \\

    MMDiT                                & Flow Matching & 6.05 \\
    MMDiT+REPA                           & Flow Matching & \underline{4.73} \\
    MMDiT+MeanFlow                       & Flow Matching & 5.02 \\
    \textbf{MMDiT+ BiFM} (ours)          & Flow Matching & \textbf{4.57} \\    
    \arrayrulecolor{black}\bottomrule
    
    \end{tabular}
    \captionsetup{width=.95\linewidth}
    \vspace{-0.3em}
    \caption{{\bf MSCOCO-256 Text-to-Image Generation.} FID computed from 40K samples is reported.}
    \label{tab:t2i}
    
    \vspace{-15pt}
\end{table}
\begin{table}[h!]
    \centering\small
    \begin{tabular}{llll}
        \toprule
        {\bf Setting}  & {\bf Model} & {\bf FID$\downarrow$} & {\bf NFE}\\
        
        \midrule
        \multirow{3}{*}{Multi-Step} 
        & DDIM~\citep{song21ddim}            & 4.67       & 50 \\
        & Flow Matching~\citep{lipman2023fm} & \underline{2.63}       & 50 \\
        & \textbf{BiFM} (ours)               & {\bf 2.17} & 50 \\
    
        \midrule
    
        \multirow{4}{*}{One-Step}   
        & Rectified Flow~\citep{liu2022rectifiedflow} & 4.85          & 1 \\
        & sCT~\citep{lu2024scd}                       & \underline{2.85}          & 1 \\
        & MeanFlow~\citep{geng2025meanflow}           & 2.92          & 1 \\
        & \textbf{BiFM} (ours)                         & \textbf{2.75} & 1 \\
        
        \bottomrule
    
    \end{tabular}
    \captionsetup{width=.95\linewidth}
    \vspace{-0.7em}
    \caption{{\bf Unconditional CIFAR-10 Results.} We include both multi-step and one-step FID.}
    \label{tab:img_small}

\end{table}


\noindent{\bf Few-Step Generation.} For the CIFAR-10 setting, we use the Flow Matching~\citep{lipman2023fm} configuration. The model is a U-Net backbone trained from scratch for 500 epochs,
Table~\ref{tab:img_small} shows quantitative results.  On CIFAR-10, BiFM improves both multi-step and one-step FID. With 50 NFE, BiFM reaches FID $2.17$, improving over Flow Matching. With 1 NFE, BiFM sets the best FID on CIFAR-10 ($2.75$ vs.\ $2.85$ for sCT and $2.92$ for MeanFlow).



\noindent{\bf ImageNet-256 training from scratch.}
In \Cref{tab:scale_train}, when training vanilla SiT variants from scratch, BiFM consistently lowers FID across model scales (e.g., SiT-XL/2: $17.2 \rightarrow 15.5$), indicating benefits beyond fine-tuning.


\begin{table}[h!]
    \centering\small
    \resizebox{0.6\linewidth}{!}{%
    \begin{tabular}{l l l}
        \toprule
         \textbf{Model}  & \textbf{Epochs} & \textbf{FID$\downarrow$}  \\
         \midrule
         SiT-B/2  & 80 & 33.0 \\
         {SiT-B/2\bf\footnotesize + BiFM}  & 80  & \textbf{27.8} \\
         \arrayrulecolor{black!40}\midrule
         SiT-L/2 & 80 & 18.8 \\
         {SiT-L/2\bf\footnotesize + BiFM}  & 80  & \textbf{16.4} \\
         \arrayrulecolor{black!40}\midrule
         SiT-XL/2  & 80   & 17.2\\
         {SiT-XL/2\bf\footnotesize + BiFM}  & 80  & \textbf{15.5} \\
        \arrayrulecolor{black}\bottomrule
    \end{tabular}
    }
    {
    \captionsetup{width=.95\linewidth}
    \caption{{\bf ImageNet Performance across Model Size.} Under multi-step sampling, BiFM training also consistently improves performance across model sizes.}
    \label{tab:scale_train}
    }
    
    \vspace{-16pt}
\end{table}

\subsection{Ablation Studies}
\label{subsec:ablation}

We ablate key design choices in 1-NFE ImageNet-256 training by varying one component at a time while keeping the architecture and training budget fixed; \Cref{tab:ablation} summarizes the effect on FID and marks our default settings.

\noindent{\bf Time-interval conditioning.} Conditioning on $(t, t' - t)$ yields the best FID (55.22) versus $(t, t')$ (59.37), $(t'-t)$ alone (60.86), or adding a discrete direction flag (69.01). Interpreting $(t' - t)$ as an explicit interval length helps the network to model the integrated flow over variable spans—directly matching our average-velocity target.

\noindent{\bf Sampling of $(t, t')$.} Among uniform and log-normal samplers, the best-performing setting skews toward shorter intervals while retaining coverage of longer hops. Practically, biasing samples to smaller $|t'-t|$ stabilizes training early.

\noindent{\bf Consistency weighting $w(t, t')$.} We found a warm-up profile avoids over-regularizing at initialization and encourages bidirectional agreement as predictions sharpen.

\noindent{\bf Loss norm $p.$} Transitioning from pure L2 $(p=0)$ to a robust loss (e.g., p$\approx$0.5) improves stability and FID by soft-clipping large residuals from difficult intervals.
\vspace{-6pt}
\section{Conclusion}
\label{sec:conclusion}
\vspace{-4pt}
In this work, we introduce BiFM, a novel framework for jointly learning few-step image generation and inversion within a single model. BiFM extends flow matching continuous time-interval supervision to both time directions to deliver accurate few-step editing. We validate its effectiveness on inversion-based image editing and generation tasks, where BiFM consistently outperforms baselines. We also conduct ablation studies on 1-NFE image generation to justify our design choices.

\section*{Acknowledgments}

\noindent We thank reviewers and AC for their time and effort in reviewing our submission.
This research is funded in part by an ARC (Australian Research Council) Discovery Grant of DP220100800. Prof.~Hongdong Li holds concurrent appointments as a Full Professor with the ANU and as an Amazon Scholar with Amzon (part time). This paper describes work performed at ANU and is not associated with Amazon.

{
    \small
    \bibliographystyle{ieeenat_fullname}
    \bibliography{main}
}

\clearpage


\setcounter{figure}{0}
\setcounter{section}{0}
\setcounter{table}{0}

\renewcommand{\thesection}{\Alph{section}} 
\renewcommand{\thetable}{\Alph{table}}
\renewcommand{\thefigure}{\Alph{figure}}






\maketitlesupplementary

\section{Extensive Backgrounds}

\paragraph{Denoising Diffusion Models} generate images from noise by learning a reverse process of a predefined forward diffusion process. The forward diffusion process is formulated as a Markov process starting from data space to a prior noise space after multiple time steps. Specifically, given the number of discrete timesteps $T$, mean schedule $\{\alpha_t\}_{t=1}^{T}$, and variance schedule $\{\sigma^2_{t}\}_{t=1}^T$, the forward process is formalized as \cref{eq:forward}, where $\rvx_0\sim p_\text{data}(\rvx)$ and $\rvx_T\sim \mathcal{N}(0,I)$. The learned reverse process of \cref{eq:forward} is \cref{eq:reverse}:
\begin{gather}
    q(\rvx_t|\rvx_0) = \mathcal{N}(\rvx_t; \alpha_t\rvx_0, \sigma_t^2 I) \label{eq:forward} \\
    p_{\theta}(\rvx_0)=\int p(\rvx_T)\prod_{t=1}^{T}p_{\theta}(\rvx_{t-1}|\rvx_t) \label{eq:reverse}
\end{gather}
The log-likelihood of samples from denoising diffusion models can be decomposed as:
\begin{align}
    \mathbb{E}\left[-\log p_\theta(\rvx_0)\right] \leq &\mathbb{E}_q\left[ - \sum_{t\geq 1} \log\frac{p_\theta(\rvx_{t-1}|\rvx_{t})}{q(\rvx_t|\rvx_{t-1})} \right] \notag \\
    &-\log p(\rvx_T) \label{eq:diffusion_likelihood}
\end{align}
By considering only optimization terms associated with the learned network (expressed as $\epsilon_\theta$), the training objective can be expressed by ELBO of \cref{eq:diffusion_likelihood} as \cref{eq:obj_simple}, a noise-prediction parameterization found effective by \citet{ho2020ddpm}:
\begin{equation}\label{eq:obj_simple}
    \mathcal{L}=\mathbb{E}_{t,\rvx_0,\epsilon}\|\epsilon-\epsilon_\theta(\alpha_t \rvx_0+\sigma_t \epsilon, t)\|^2
\end{equation}
\paragraph{Flow Matching} constructs a time dependent path which transports noise distribution $\rvx_0\sim \mathcal{N}(\rvx;0,I)$ into data distribution $\rvx_1\sim p_{\text{data}}(\rvx)$. The transportation is described as the following flow matching ODE:
\begin{gather}
    \frac{d}{dt}\phi_t(\rvx) = v_t(\phi_t(\rvx)), \\
    \rvx_t = \phi_t(\rvx_0) , \phi_0(\rvx) = \rvx
\end{gather}
A flow model is uniquely determined by its learned velocity field $v_\theta(\rvx_t,t)$. Flow matching modifies from the noise prediction in denoising diffusion trajectory to velocity prediction in probability distribution transport flow, which simplifies the overall framework.
A practical flow matching training objective, conditional Flow Matching Loss (CFM), can be written as:
\begin{equation}\label{eq:cfm}
    \mathcal{L}_{\text{CFM}}=\mathbb{E}_{t,\rvx_0,\rvx_1}|| v_\theta(\rvx_t,t)-v_t(\rvx_t | \rvx_0, \rvx_1) ||^2
\end{equation}
where the target velocity $v_t$ is the conditional velocity. where $v_t | \rvx_0, \rvx_1 :=\rvx_0-\rvx_1$ is the per-sample velocity of the flow, $v_{\theta}(\rvx_t,t)$ is the velocity prediction from the leaned neural network $\theta$.

\section{Additional Implementation Details}

\paragraph{Model Configuration.}
For image editing experiments, we adopt Stable Diffusion 3 Medium (SD3-M)~\citep{esser2024sd3}, a Multimodal Diffusion Transformer (MMDiT) operating in the latent space of its VAE, conditioned on three pretrained text encoders: CLIP-L/14, CLIP-G/14, and T5-XXL, following the official SD3 design. Following \citep{chadebec2025flash}, we train LoRA adapters only in the MMDiT blocks. In addition, we introduce a trainable extra time-interval embedding module that augments the SD3 timestep conditioning, which has the same architecture as the time embedding in original SD3, and is zero-initialized. The total trainable parameters are LoRA weights injected into attention/ MLP projections, and extra time-embedding parameters.
For SD3 experiments, we use 32 H100 GPUs for fine-tuning;  100 epochs takes $\sim120$ hours. 

\paragraph{Dataset Configuration.}
For training \textsc{BIFM} on a pretrained Stable Diffusion 3 model~\citep{esser2024sd3}, we utilized MagicBrush dataset~\citep{zhang2023magicbrush} with 10K manually annotated real image editing triplets. We generate captions for source and target images using BLIP-2. We train our model with batch size of 4 and learning rate $1e^{-5}$ with Adam optimizer.

\paragraph{Training Configuration.} We do not train BiFM with CFG guidance (unlike MeanFlow configurations) to preserve sampling flexibility across guidance values. We train without guidance and apply CFG only at inference when appropriate. 
For T2I results we use CFG scale 4. For ImageNet results we do not apply CFG. ImageNet training takes 80 epochs ($\sim150k$ steps, batch size 256), and Table~\ref{tab:re_gen}(b) shows BiFM achieves noticeable gains over MeanFlow after 80 epochs.

\section{Additional Experimental Details}

\paragraph{Sampling and Evaluation Details.} In Figure~\ref{fig:edit_visual}, NFE/steps used for each method are: PnP Inv 50, RF-Edit 30, FlowEdit 28, ReNoise 4, SwiftEdit 1, and BiFM 1. 

\figTrainingBenefitFID

We show that BiFM offers benefit in image generation training.~\cref{fig:cifarbenefit} presents training curves of FID versus epoch for baseline (FM) and FM augmented with BiFM. Across the entire training trajectory, \textsc{FM+BiFM} achieves consistently lower FID than FM alone, indicating faster convergence and better generative quality.

We include additional baselines for image editing and generation experiments (see Table~\ref{tab:re_edit} and~\ref{tab:re_gen}).
As shown in Table~\ref{tab:re_edit}, BiFM achieves better background preservation than DNAEdit, while performing a bit worse on CLIP semantics.
In the Few-Step regime, BiFM attains higher SSIM/PSNR and higher CLIP semantics than InstantEdit (4 NFE) and FireFlow, while trading off some LPIPS.
We include EditFT, InstantEdit, and FlowEdit as baselines in Table~\ref{tab:re_edit}. We also match methods by NFE (e.g., 1-step/4-step/multi-step) to align with the few-step focus of BiFM.

\begin{table*}[htbp]
    \centering
    \small
    \resizebox{\linewidth}{!}{%
    \begin{tabular}{lllllllll}
        \toprule
        \multirow{2}{*}{\textbf{Settings}}  & \multirow{2}{*}{\textbf{Methods}} & \multirow{2}{*}{\textbf{Steps}} & \multicolumn{4}{c}{\textbf{Background Preservation}} & \multicolumn{2}{c}{\textbf{CLIP Semantics}} \\
        \cmidrule(lr){4-7} \cmidrule(lr){8-9}
        & & & {\bf LPIPS}$_{\times10^3}\downarrow$ & {\bf SSIM}$_\%\uparrow$ & {\bf PSNR}$\uparrow$  & {\bf MSE}$_{\times10^4}\downarrow$  & {\bf Whole}$\uparrow$ & {\bf Edited}$\uparrow$ \\ 
        
        \midrule
        \multirow{4}{*}{Multi-Step} 
        & DNAEdit           & 28 & 112.60 & 83.69 & 23.24 & 67.32 & {\bf 28.90} & \underline{23.66} \\
        & FlowEdit         & 28 & 112.19 & 83.08 & 21.96 & 94.99 & 25.25 & 22.58 \\
        & EditFT            & 30 & \underline{80.55}  & {\bf 91.50} & \underline{26.62} & {\bf 40.24} & {25.74} & 22.27 \\
        
        & {\bf BiFM} (ours) & 50 & {\bf 47.01} & \underline{87.50} & {\bf 29.89} & \underline{42.66}  & \underline{27.42} & {\bf 24.66} \\
            
        \midrule
        \multirow{4}{*}{Few-Step}  
        & FireFlow          & 18 & -     & 82.49 & 23.03 & -     & 26.02 & 22.81 \\
        & InstantEdit       & 12 & {\bf 35.27} & {\bf 87.40} & {\bf 29.63} & {\bf 24.57} & 26.06 & 22.73 \\
        & InstantEdit       & 4  & \underline{44.39} & 86.44 & 27.96 & \underline{34.94} & \underline{26.28} & \underline{22.82} \\
        
        & {\bf BiFM} (ours) & 4  & 67.25 & \underline{87.29} & \underline{28.92} & 54.92 & {\bf 26.77} & {\bf 23.58} \\
        

        
        \bottomrule
    \end{tabular}
    }
    {
    \captionsetup{width=\linewidth}
    \vspace{-0.3em}
    \caption{\small {\bf Additional PIE-Bench Image Editing Performance.} }
    \label{tab:re_edit}
    }
    
\end{table*}
\begin{table}[h!]
    \centering
    \small 
    \begin{subtable}[t]{0.45\columnwidth}
        \centering
        \begin{tabular}{l l l}
            \toprule
            {\bf Method} & {\bf Model} & {\bf FID$\downarrow$} \\    
            \midrule
            FM            & MMDiT & 6.05 \\
            REPA          & MMDiT & \underline{4.73} \\
            MeanFlow      & MMDiT & 5.02 \\
            \textbf{BiFM} & MMDiT & \textbf{4.57} \\        
            \arrayrulecolor{black}\bottomrule
        \end{tabular}
        \label{tab:re_t2i}
        \caption{Text-to-image generation result on MSCOCO. We re-implement MeanFlow on MMDiT for results.}
    \end{subtable}
    \hfill
    \begin{subtable}[t]{0.45\columnwidth}
        \centering
        \begin{tabular}{l l l}
            \toprule
            {\bf Model}  & {\bf Method} & {\bf FID$\downarrow$}  \\
            \midrule
            SiT-B/2 & MeanFlow   & 28.2 \\
            SiT-B/2 & {\bf BiFM}  & {27.8} \\
            SiT-L/2 & MeanFlow  & \underline{17.0} \\
            SiT-L/2 & {\bf BiFM}  & {\bf 16.4} \\
            \arrayrulecolor{black}\bottomrule
        \end{tabular}
        \label{tab:re_imgnet}
        \caption{ImageNet-256 generation. We do not distill / apply CFG for this experiment.}
    \end{subtable}
    \caption{{\bf Image Generation Results.} NFE=50.}
    \label{tab:re_gen}   
\end{table}

In Table~\ref{tab:re_gen}, we add MeanFlow results on MSCOCO and MeanFlow training from scratch on ImageNet, to validate improvements beyond CIFAR-10.

\begin{figure}[t]
    \centering
    \setlength{\tabcolsep}{1.5pt}
    \begin{tikzpicture}
    \node[inner sep=0pt] (table) {%
    \begin{tabular}{l c c c}

        {\bf Method} & {\bf 1-step} & {\bf 4-steps} & {\bf 28-Steps} \\

        \raisebox{0.06\textwidth}{\rotatebox[origin=t]{0}{\scalebox{1.0}{\begin{tabular}{l@{}l@{}l@{}} \bf{FlowEdit} \end{tabular}}}} & 
        \includegraphics[width=0.12\textwidth,height=0.12\textwidth]{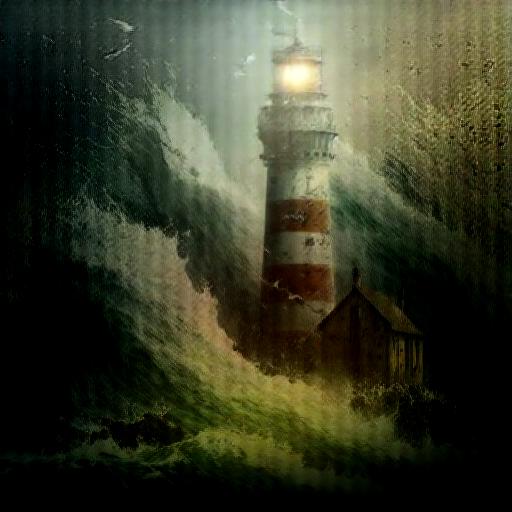} &
        \includegraphics[width=0.12\textwidth,height=0.12\textwidth]{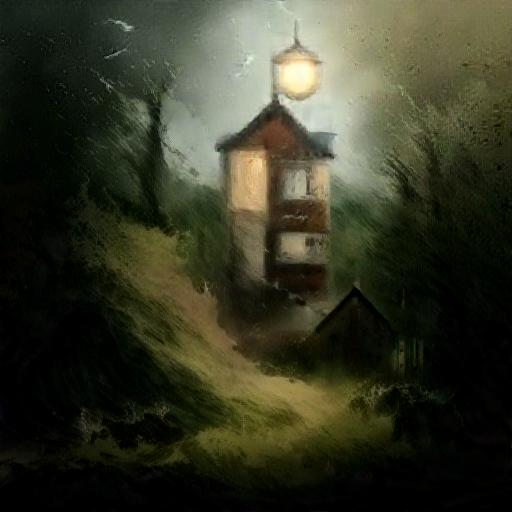} &
        \includegraphics[width=0.12\textwidth,height=0.12\textwidth]{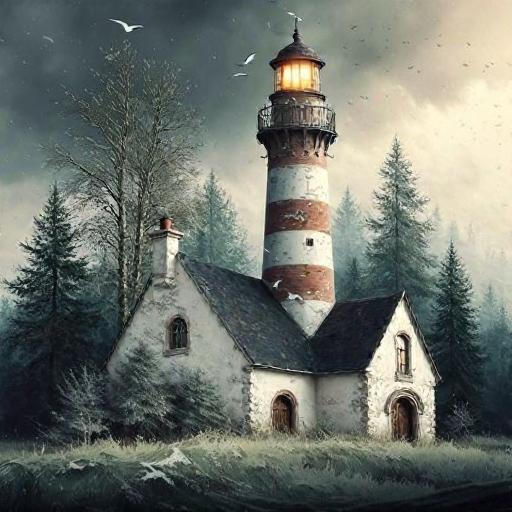} \\ 

        \raisebox{0.06\textwidth}{\rotatebox[origin=t]{0}{\scalebox{1.0}{\begin{tabular}{l@{}l@{}l@{}} \bf{BiFM} \end{tabular}}}} &
        \includegraphics[width=0.12\textwidth,height=0.12\textwidth]{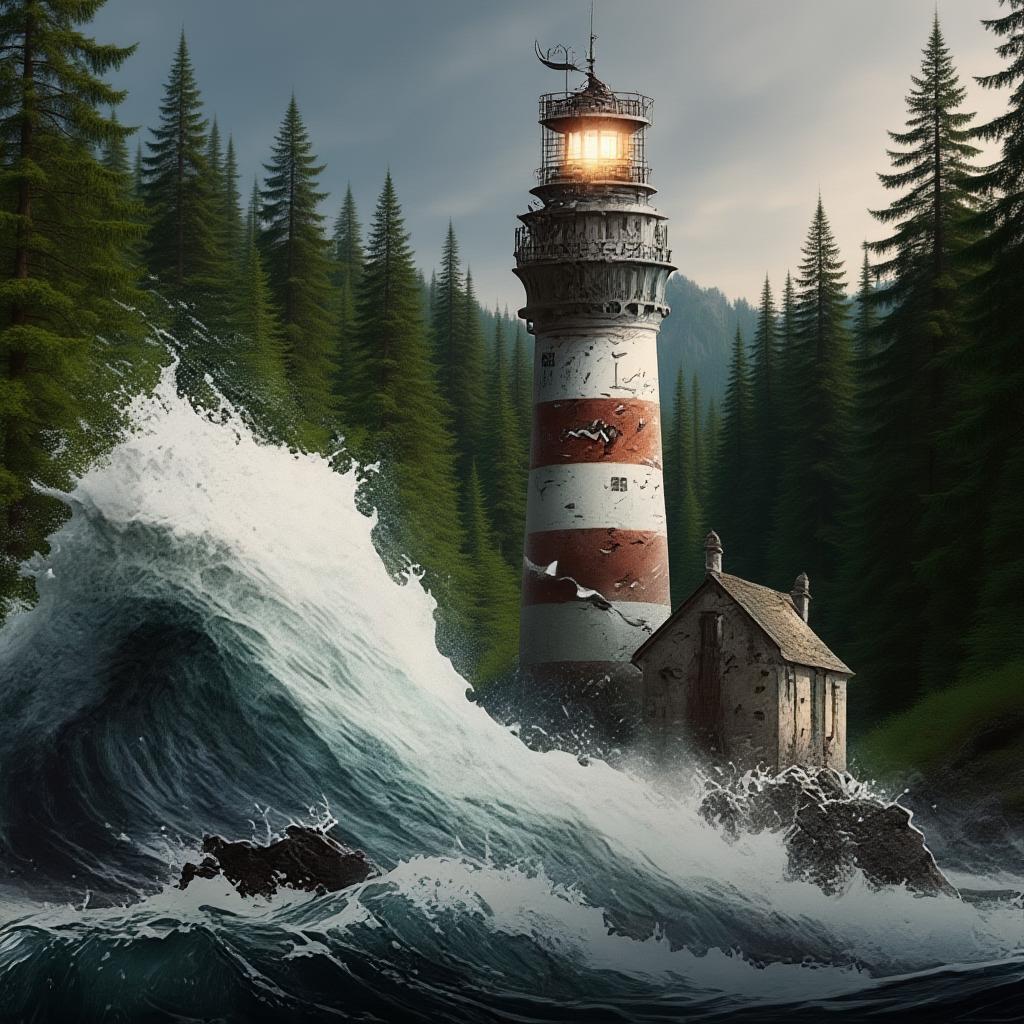} &
        \includegraphics[width=0.12\textwidth,height=0.12\textwidth]{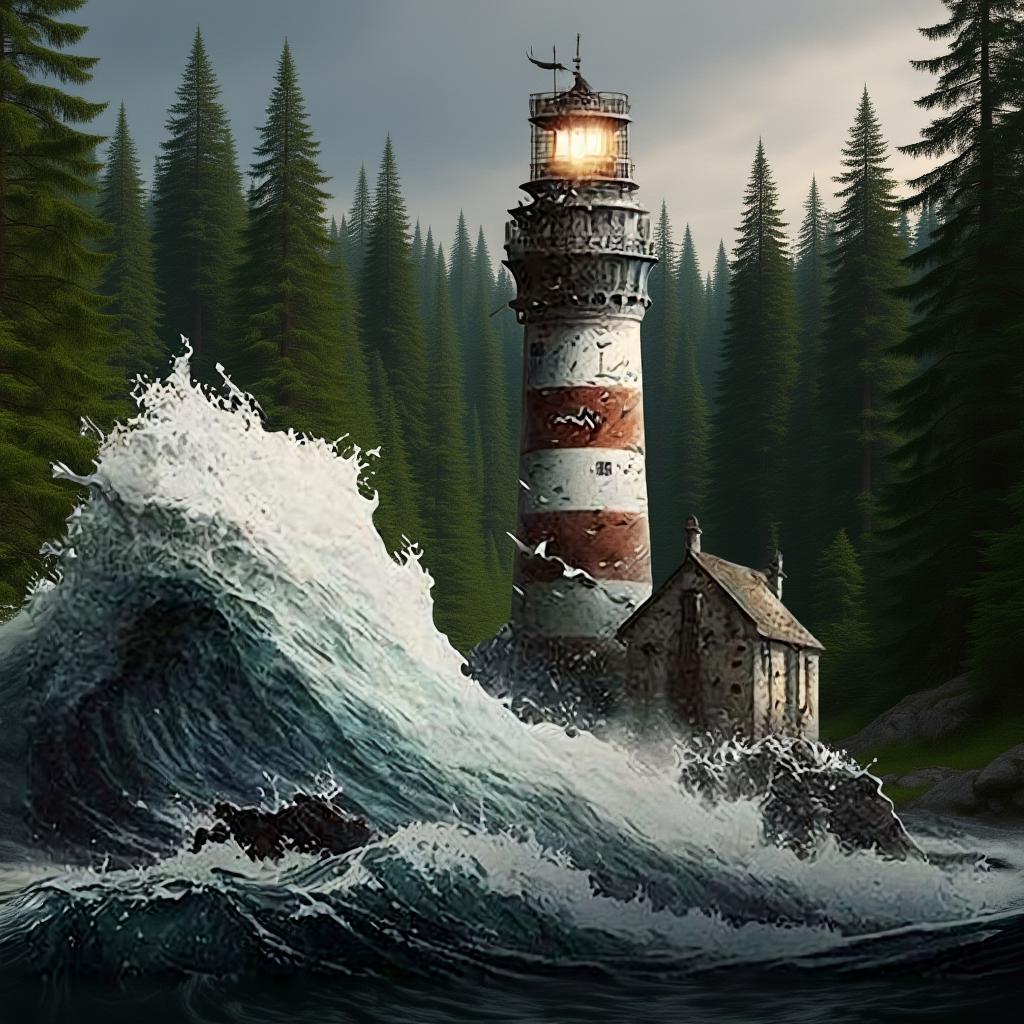} &
        \includegraphics[width=0.12\textwidth,height=0.12\textwidth]{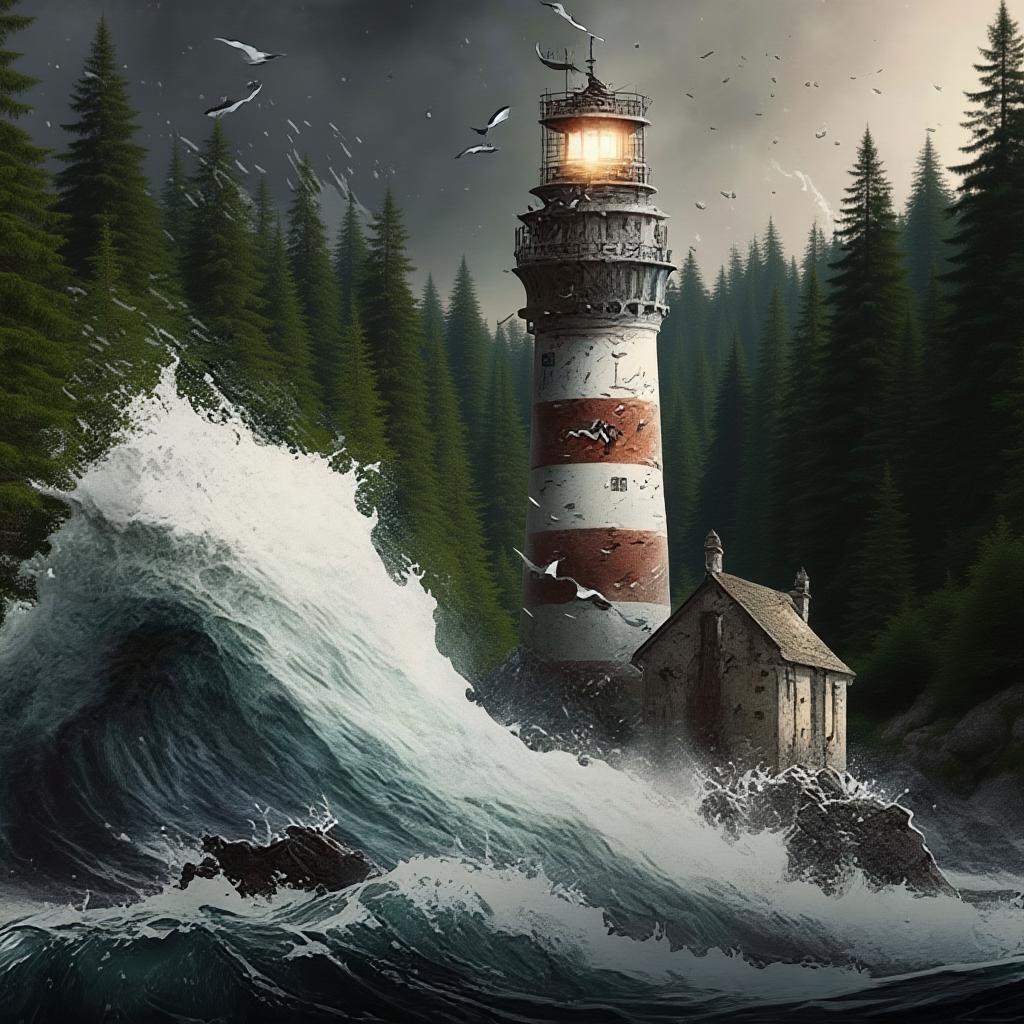} \\ 

    \end{tabular}
    };
    \end{tikzpicture}
    \caption{\small {\bf Image Editing Visualization.}}
    \label{fig:re_visual}
\end{figure}

\section{More Generation Visualization}

\begin{figure*}[h]
    \centering
    \includegraphics[width=0.75\linewidth]{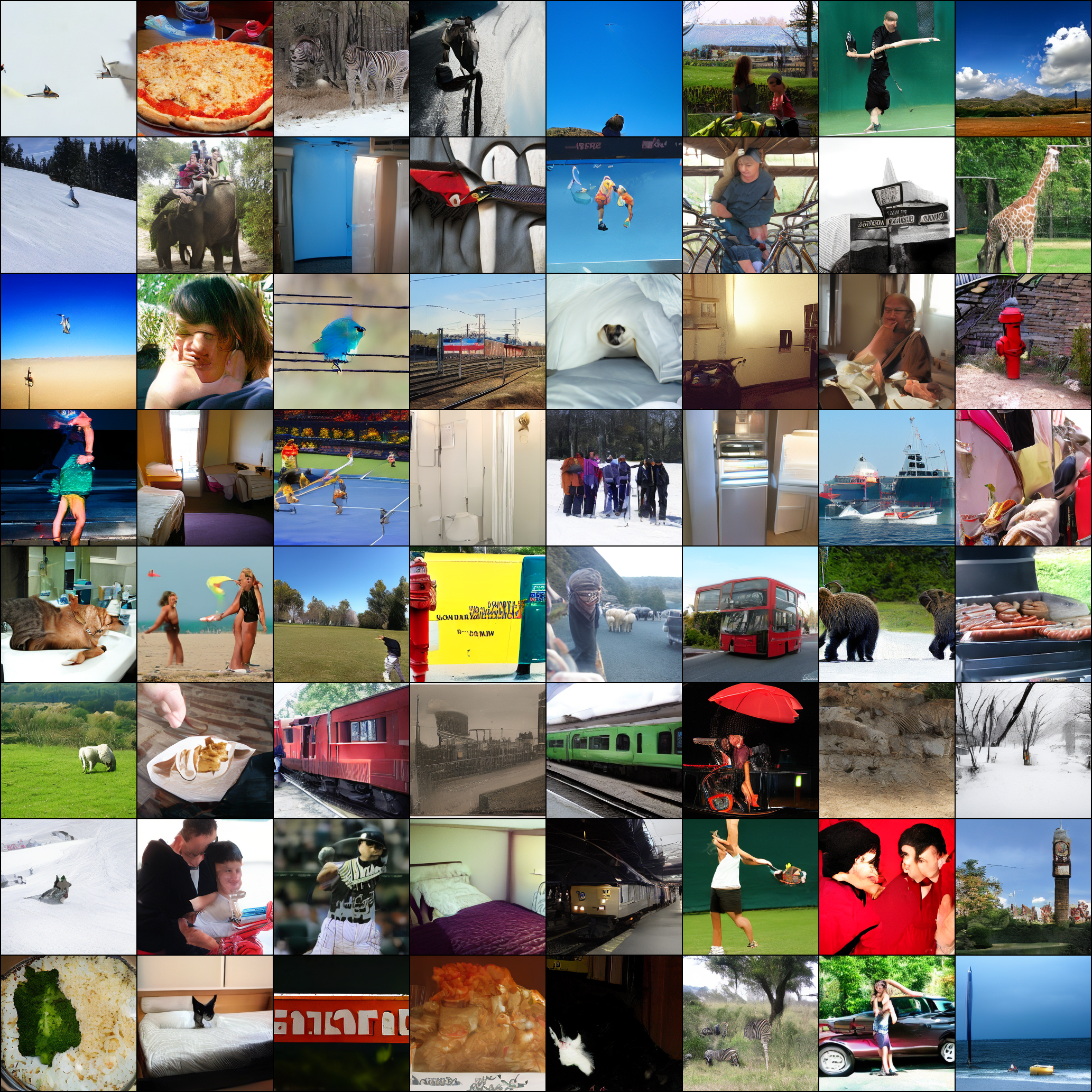}
    \caption{\small {\bf MSCOCO-256 Text-to-Image Generation Visualization Results. } Model trained for 100K iterations.}
    \label{fig:mscoco_samples}
\end{figure*}

\begin{figure*}[h]
  \centering
  \begin{subfigure}{0.5\textwidth}
    \includegraphics[width=\linewidth]{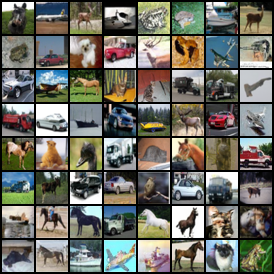}
  \end{subfigure}
  \begin{subfigure}{0.5\textwidth}
    \includegraphics[width=\linewidth]{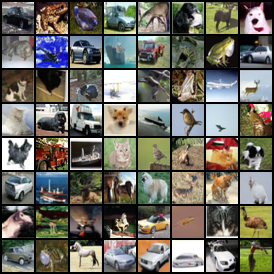}
  \end{subfigure}
  \caption{\small {\bf CIFAR-10 Generation Visualization Results. } Model trained for 500 epochs.}
  \label{fig:cifar_samples}
\end{figure*}

\begin{figure*}[h]
  \centering
  \begin{subfigure}{0.5\textwidth}
    \includegraphics[width=\linewidth]{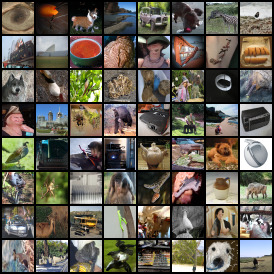}
  \end{subfigure}
  \begin{subfigure}{0.5\textwidth}
    \includegraphics[width=\linewidth]{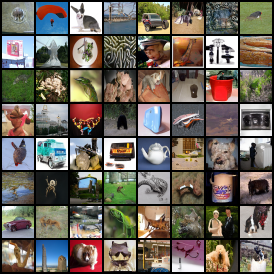}
  \end{subfigure}
  \caption{\small {\bf ImageNet-32 Generation Visualization Results.} Model trained for 80 epochs.}
  \label{fig:imgnet_samples}
\end{figure*}

In this section we provide more visualization samples from image generation experiments.
In Figure~\ref{fig:re_visual}, we show editing examples comparing BiFM and FlowEdit under 1-step, 4-step and 28-step settings.
\cref{fig:mscoco_samples} shows uncurated samples generated by vanilla MMDiT using random prompts from MSCOCO-256 dataset. For small-resolution datasets, \cref{fig:cifar_samples} and \cref{fig:imgnet_samples} display uncurated 32$\times$32 samples from CIFAR-10 and ImageNet-32, respectively, generated by a U-Net model trained with BiFM.

\end{document}